\documentclass[pdflatex,sn-mathphys-num]{sn-jnl}

\usepackage{graphicx}%
\usepackage{multirow}%
\usepackage{amsmath,amssymb,amsfonts}%
\usepackage{amsthm}%
\usepackage{mathrsfs}%
\usepackage[title]{appendix}%
\usepackage{xcolor}%
\usepackage{textcomp}%
\usepackage{manyfoot}%
\usepackage{booktabs}%
\usepackage{algorithm}%
\usepackage{algorithmicx}%
\usepackage{algpseudocode}%
\usepackage{listings}%
\usepackage{comment}
\usepackage{multirow}
\usepackage{tabularx}
\usepackage[table]{xcolor}
\usepackage{subcaption}
\usepackage{array}
\usepackage{cleveref}

\usepackage[font=footnotesize]{caption}
\usepackage{enumitem}
\usepackage{adjustbox}
\usepackage{changepage}
\usepackage{rotating} 

\usepackage{amsmath, amssymb, amsthm, bm}
\usepackage{mathtools}
\usepackage{hyperref}

\usepackage{placeins}

\newcolumntype{C}[1]{>{\centering\arraybackslash}p{#1}}

\theoremstyle{thmstyleone}%
\theoremstyle{thmstyletwo}%

\theoremstyle{thmstylethree}%

\begin{document}

\title[Article Title]{

Knowledge-Inclusive Adaptive Physics-Informed Neural Network for Microbial Interaction Modelling}


\author[1]{\fnm{Ravisha} \sur{Rupasinghe}}

\author[1]{\fnm{Rajith} \sur{Vidanaarachchi}}
\equalcont{These authors contributed equally to this work.}

\author[1]{\fnm{Asela} \sur{Hevapathige}}
\equalcont{These authors contributed equally to this work.}

\author[1]{\fnm{Sachith} \sur{Seneviratne}}

\author[2]{\fnm{Sen-Lin} \sur{Tang}}

\author*[1]{\fnm{Saman} \sur{Halgamuge}}\email{saman@unimelb.edu.au}

\affil[1]{\orgdiv{AI, Optimisation and Pattern Recognition Lab, Department of
Mechanical Engineering}, \orgname{University of Melbourne}, \orgaddress{\city{Melbourne}, \state{Victoria}, \country{Australia}}}

\affil[2]{\orgdiv{Biodiversity Research Center}, \orgname{Academia Sinica}, \orgaddress{\city{Taipei}, \country{Taiwan}}}

\abstract{Physics-Informed Neural Network (PINN) is a way of including knowledge in the form of equations in Machine Learning methods. Beyond equations, knowledge exists in other forms, such as text and network structure.
While existing PINN-based approaches discover equation parameters from data, they rely solely on experimental measurements. We propose a new PINN framework that enriches parameter discovery by incorporating auxiliary knowledge sources. 
We instantiate our framework for microbiology, where generalised Lotka-Volterra (gLV) serves as a biological foundation for modelling microbial communities. 
We demonstrate that incorporating knowledge improves microbial community modelling.
Our framework enriches the gLV parameters using peer-reviewed metagenomics literature, as text provides biological context on external influences that gLV alone cannot capture. We combine this knowledge with experimental measurements of microbial abundance using a data-driven integration approach.
We integrate network-based structural knowledge by explicitly modelling microbial interactions. 
Our knowledge-inclusive framework infers microbial networks, revealing ecological insights. We validate these findings against ecological roles documented in the literature.
We evaluate on real and simulated datasets spanning human- and plant-associated microbial communities. Our framework improves over the state-of-the-art by up to 53\%, even without knowledge. Knowledge addition yields gains of up to 23\% in Bray-Curtis Dissimilarity-based accuracy and 47\% in $\text{R}^2$.

}




\maketitle

\section{Introduction}\label{sec1}

Physics-Informed Neural Networks (PINNs) \cite{raissi2019physics} are a prominent modelling framework, excelling at data-driven solutions for systems governed by physical laws. A particularly important direction within PINNs is adaptive parameter learning, where model parameters are inferred from experimental data rather than fixed a priori \cite{raissi2019physics, fontanarrosa2025exploring}. However, experimental data captures only a narrow slice of system behaviour under specific controlled conditions, while the same system may respond very differently across varying contexts, environments, and regimes that no single experiment can cover. The broader heterogeneous knowledge accumulated across the scientific literature, curated databases, and domain expertise encodes precisely this wider contextual understanding. No existing approach attempts to learn physics equation parameters guided by these broader knowledge sources, leaving a critical gap between what the equations encode and what is collectively known about a system. Knowledge-Inclusive Machine Learning (KIML) \cite{kiml} provides this kind of framework, combining experimental data with such heterogeneous knowledge sources within machine learning architectures so that domain knowledge and empirical observations jointly inform the learning process.

\begin{figure}[h]
\centering
\includegraphics[width=1.0\textwidth]{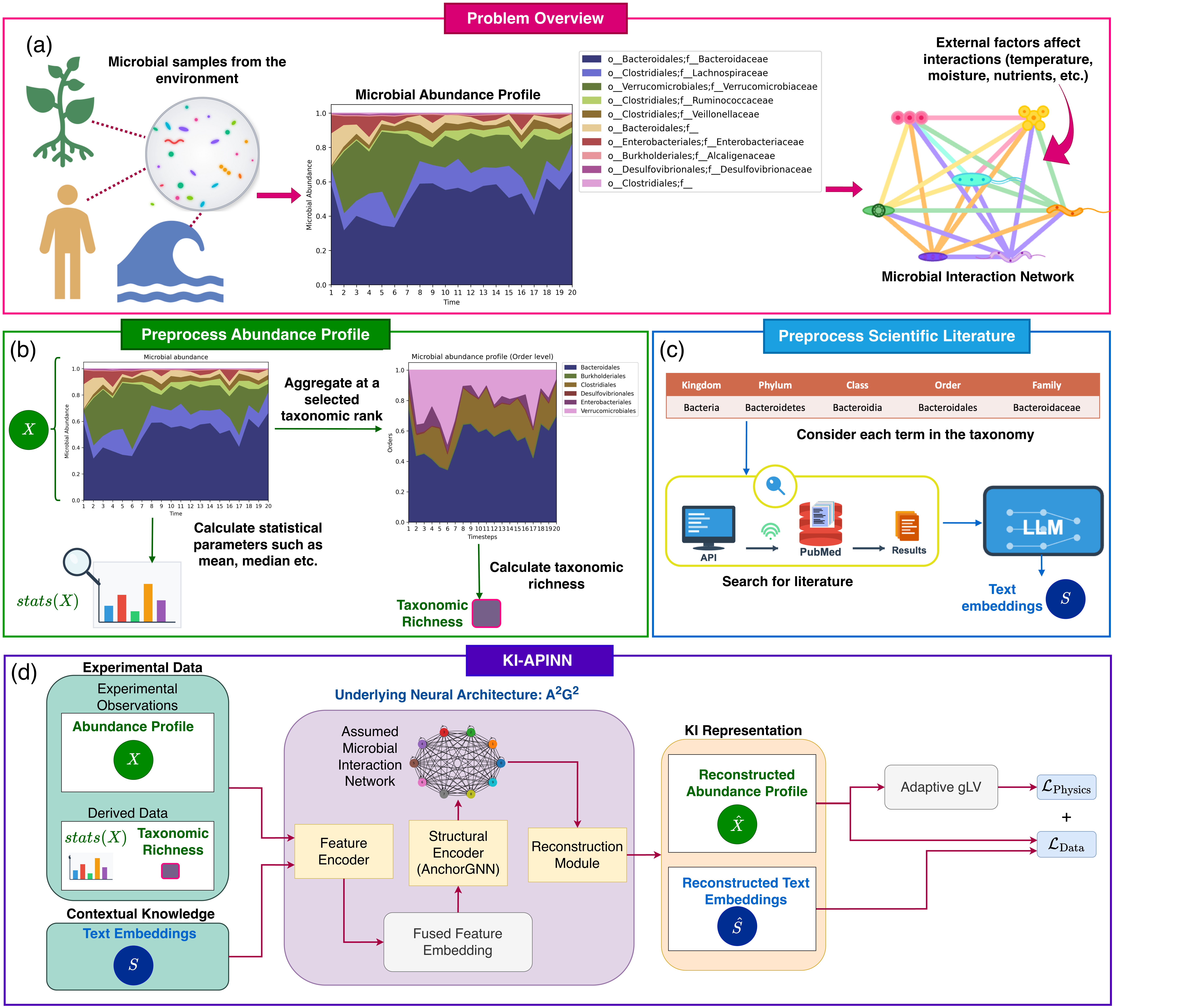}
\caption{Knowledge Inclusive ML (KIML) approach. \textbf{(a)} Understanding microbiome function requires characterising microbial interactions shaped by both microbial abundance and environmental factors including temperature, nutrients, etc.
\textbf{(b)} The microbial abundance profile is preprocessed to obtain both the statistical parameters of the abundance profile and the taxonomic richness derived by considering a specific taxonomy level in the data. Both these data types, along with the abundance profile, are fed as inputs to the ML model.
\textbf{(c)} Contextual knowledge in the form of text information in selected PubMed literature is preprocessed to create text embeddings using a Language Model.
\textbf{(d)} Knowledge Inclusive Adaptive PINN (KI-APINN) for microbial interaction modelling. Both experimental data and contextual knowledge are given as input to the model. The underlying neural architecture is the \textbf{A$^2$G$^2$} (\textbf{A}ttention-based feature encoding with \textbf{A}nchor\textbf{G}NN for microbial network topology and \textbf{G}ated Recurrent Unit (\textbf{G}RU) for capturing temporal dynamics). A$^2$G$^2$ has three main components: the feature encoder---creates a fused feature embedding using an attention mechanism on all forms of input data, the structural encoder (AnchorGNN)---models the microbial network, and the reconstruction module---reconstructs a Knowledge Inclusive (KI) representation. The pairwise interactions and growth rates in the generalised Lotka–Volterra (gLV) equation are learned through the KI-APINN.
}\label{fig:pipeline}
\end{figure}

Microbial interaction modelling is a domain where this gap is particularly important. Microbes are the most abundant life forms on Earth, shaping the chemical and physical properties of every environment they inhabit \cite{blaser2016toward, cani2019microbial}. Their interactions within communities are central to host health, ecosystem function, and biotechnological applications \cite{clemente2012impact, cho2024novel}, yet characterising those interactions remains a fundamental open problem \cite{bergey}. Microbial dynamics are governed by both internal pairwise interactions among taxa and external environmental influences \cite{faust2012microbial, baranwal2022recurrent} such as temperature, nutrient availability, and treatment effects (Fig. \ref{fig:pipeline}a). Critically, microbial communities follow known governing equations with biologically meaningful parameters, making pure algorithmic optimisation insufficient since a solution that fits the data without respecting the underlying biological structure cannot be trusted or interpreted. The generalised Lotka-Volterra (gLV) equation is the standard biological foundation for microbial interaction modelling \cite{imparo}, encoding pairwise interaction dynamics and growth rates as explicit parameters, and a PINN framing is therefore the natural choice since it preserves this biological structure while enabling data-driven parameter learning. Furthermore, abundance profiles record who is present and in what quantity but carry no information about the environmental conditions shaping those dynamics, and real microbial communities respond to dozens of simultaneous influences that no single experiment can fully characterise. Heterogeneous knowledge sources, such as peer-reviewed metagenomics literature, accumulated across thousands of studies spanning diverse environments and conditions, as well as network interaction information between microbes, encode precisely this broader contextual knowledge, making it a natural and powerful complement to any single experimental dataset.

A few methods have been proposed for microbial interaction modelling \cite{imparo, fontanarrosa2025exploring, baranwal2022recurrent, ruaud2024modelling}. Long Short-Term Memory (LSTM) based approaches \cite{baranwal2022recurrent} attempt to capture temporal dynamics from abundance profiles but construct interaction networks post-hoc without any topological or mechanistic grounding, which can produce misleading representations of community structure. Graph Neural Network (GNN) based approaches \cite{ruaud2024modelling} instead model the topology of microbial interactions directly but lack any representation of temporal dynamics. While gLV-based methods retain the biological equation structure, mostly estimate the parameters from the abundance data alone \cite{imparo}, ignoring the heterogeneous knowledge sources that could make those parameters more reliable and context-aware. No single method captures temporal dynamics, interaction topology, and contextual knowledge simultaneously, leaving the complete picture of microbial community behaviour out of reach (Table S1). Beyond these individual limitations, all existing methods rely solely on abundance data, cannot incorporate the rich contextual knowledge available in the metagenomics literature, and have been evaluated only on a narrow set of human gut microbiome datasets.

\begin{figure}[h]
\centering
\includegraphics[width=1.0\textwidth]{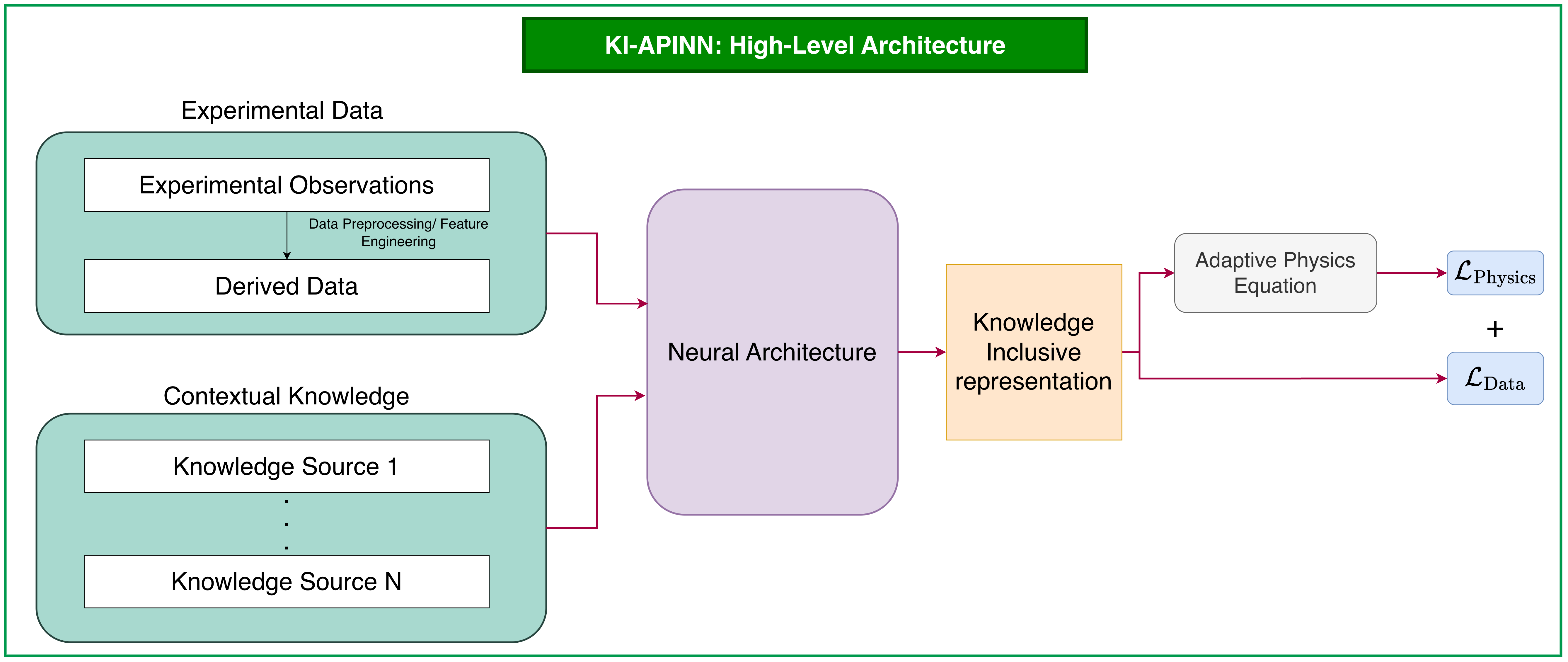}
\caption{The high-level overview of our Knowledge-Inclusive Adaptive PINN (KI-APINN). This is a model-agnostic framework that can be integrated into any neural architecture. The architecture fuses experimental inputs with multi-source contextual knowledge to create a knowledge-inclusive (KI) representation, which guides physics parameter learning in the adaptive physics equation. The total loss comprises a physics term $L_{\text{Physics}}$ and a data term $L_{\text{Data}}$ for training the neural architecture. 
}
\label{fig:ki_apinn}
\end{figure}

We address these gaps by introducing the Knowledge-Inclusive Adaptive PINN (KI-APINN) (Fig. \ref{fig:ki_apinn}), a model-agnostic framework in which physics equation parameters are not fixed scalars but learned functions of a knowledge-inclusive representation formed by fusing experimental data with auxiliary knowledge sources. This is a strictly more general formulation than standard parameter-discovery PINNs (which we term Adaptive PINNs---APINNs) \cite{raissi2019physics}, which emerge as a special case when no auxiliary knowledge is incorporated. The framework imposes no constraints on the choice of neural surrogate, fusion mechanism, or parameter network, making it applicable wherever physics equations with estimable parameters coexist with
heterogeneous knowledge sources. To demonstrate its utility concretely, we instantiate KI-APINN for microbial interaction modelling (Fig. \ref{fig:pipeline}d). We realise this instantiation through $\mathrm{A}^2\mathrm{G}^2$, which integrates three complementary components. An attention-based feature encoder fuses temporal
abundance profile, statistical parameters, taxonomic richness, and text embeddings derived from peer-reviewed metagenomics literature, with per-modality attention weights providing an interpretable measure of each knowledge source's contribution to the final prediction. A novel structural encoder (AnchorGNN) is proposed to capture the topology of microbial interactions at sub-quadratic cost by sparsifying the interaction graph around a learned set of anchor taxa. A gLV reconstruction head decodes the learned representations into per-taxon growth rates and pairwise interaction strengths, closing the loop between the neural architecture and the physics equation. While the gLV equation is not a universal description of microbial dynamics \cite{dedrick2023does, joseph2020compositional}, conditioning its parameters on the knowledge-inclusive representation allows the model to remain faithful to known biological structure while adapting to context-dependent dynamics that the equation alone cannot capture.

We consider two pipelines, each with its own assumptions and implications (Fig. \ref{fig:two_pipelines}); multi-community generalisation pipeline and the mono-community pipeline. The multi-community pipeline enables broad cross-community analysis but may dilute community-specific knowledge through aggregation. In contrast, the mono-community pipeline provides modelling of an individual community at a time, enabling integration of community-specific internal and external factors.\\

In summary, this work makes four contributions. First, we introduce KI-APINN, a model-agnostic framework that generalises parameter-discovery PINNs by conditioning physics parameters on learned knowledge-inclusive representations, with standard PINNs recovered as a formal special case when no auxiliary knowledge is present. Second, we develop $\mathrm{A}^2\mathrm{G}^2$, a concrete instantiation combining
attention-based multimodal fusion, anchor-based sparse graph encoding, and gLV-grounded parameter decoding for microbial interaction modelling. Third, we conduct the most comprehensive evaluation in this space to date, covering six datasets across human and plant microbiomes under both mono-community and multi-community generalisation pipelines. Our evaluation considers comparison against gLV-based, LSTM-based, and GNN-based baselines under both mono-community and multi-community generalisation pipelines. The proposed method improves over the state of the art by up to 53\% even without auxiliary knowledge, and knowledge incorporation yields further gains of up to 23\% in Bray-Curtis Dissimilarity (BCD) based accuracy \cite{imparo} and 47\% in $\text{R}^2$ \cite{baranwal2022recurrent, ruaud2024modelling}. Fourth, we provide biological insights into microbial interaction networks constructed from learned gLV parameters, analysing gut microbiome and ginger rhizosphere case studies where interpretable attention weights reveal which knowledge sources drive each prediction. The inferred microbial interaction networks are created from the learned interaction parameters. For the validation of the biological insights, we have utilised a Knowledge Graph (KG) from the metagenomics domain \cite{metagenomickg} that provides structured relationships between microbial taxa.

\section{Results}\label{sec:results}

This section presents the experimental datasets (Section \ref{subsec:data}), summarises the model architecture (Section \ref{subsec:model}), 
compares A$^2$G$^2$ performance with and without auxiliary (contextual) knowledge integration (Sections \ref{subsec:basic_vs_imparo}--\ref{subsec:unseen_gen}, and \ref{subsec:other_exp}), and analyses predicted microbial interactions (Sections \ref{subsec:gut_analysis}--\ref{subsec:ginger_analysis}).

\subsection{Data overview}\label{subsec:data}

To evaluate model performance, we used three datasets for the mono-community pipeline and three datasets for the multi-community pipeline. For the mono-community pipeline, we analysed adult male (M3) and female (F4) gut microbiomes \cite{caporaso}, and ginger rhizosphere microbial community data \cite{soil}. For the multi-community pipeline, we analysed \textit{in vitro} gut microbial communities \cite{clark2021design}, simulated gut microbial data \cite{baranwal2022recurrent}, and neonatal gut and respiratory microbiota \cite{neonatal}.

We used family-level taxonomic resolution for datasets in the mono-community pipeline, consistent with standard practice in microbiome studies. Family-level resolution has been widely adopted for analysing gut and rhizosphere microbiomes \cite{karwowska2024microbiome, imparo, averill2021soil} and has been shown to provide suitable resolution for machine learning tasks \cite{armour2022goldilocks}.
The taxonomic resolution for datasets used in the multi-community pipeline varied according to the nature of each dataset. For the neonatal microbiome dataset \cite{neonatal}, family-level taxonomic resolution was employed. The \textit{in vitro} gut microbiome dataset \cite{clark2021design} and simulated gut microbiome dataset \cite{baranwal2022recurrent} were analysed at species-level resolution, leveraging the precisely defined species composition of these controlled communities.

To evaluate model performance for datasets in the mono-community pipeline, we examined performance across multiple community sizes rather than at a single fixed size. Community size is defined as the number of taxa included in the analysed community. This incremental approach enabled us to observe performance variation with the sequential addition of rarer taxa. Community size was varied using three overlapping approaches: (i) selecting the top $k$ most abundant taxa \cite{imparo}, (ii) applying cumulative abundance coverage thresholds \cite{cho2024ecological}, and (iii) using a minimum relative abundance threshold \cite{reitmeier2021handling}.

For approach (i), selecting the top $k$ families reflects the natural structure of microbial abundance profiles, where taxa span a range of abundances. By selecting the top $k$ most abundant taxa for $k \in \{5, 10, 15, 20, 25, ...\}$, we progressively incorporated lower-abundance taxa into the community. For adult gut microbiomes, we considered 5-25 families. For the ginger rhizosphere, we considered 5-130 families. The extended range reflects the characteristic evenness of rhizosphere microbiome abundance distribution. For example, studies have analysed only the 10 most frequently observed bacterial groups at each taxonomic level, noting that soil microbiomes contain far more taxa than can be reasonably analysed in full \cite{averill2021soil}.

For approach (ii), the selected cumulative abundance coverage thresholds were dataset-specific. We use these thresholds to create coverage regions in the community sizes defined by (i).
For adult gut microbiomes \cite{caporaso}, we partitioned the data using 90\% and 99\% thresholds, that is, selecting families until at least 90\% or at least 99\% cumulative abundance was achieved, thereby creating three performance evaluation regions (spanning 5-25 families). For the ginger rhizosphere \cite{soil}, we used 50\% and 90\% thresholds (spanning 5-130 families). The lower threshold (50\%) accounts for the more uniform abundance distribution characteristic of rhizosphere communities compared to gut microbiomes \cite{averill2021soil}.

For approach (iii), we applied a 0.25\% minimum relative abundance threshold, established as appropriate for gut and rhizosphere microbiomes \cite{reitmeier2021handling}, ensuring it fell within the taxa range defined by approaches (i) and (ii).

The correspondence between top $k$ abundant taxa, cumulative abundance coverage, and the minimum relative abundance threshold for each dataset is provided in Tables S2 and S3. Data preprocessing details, time points considered, and community sizes for each dataset are described in Supplementary Note 4.\\

\subsection{Model overview}\label{subsec:model}

Our Knowledge-Inclusive Machine Learning (KIML) approach (Fig. \ref{fig:pipeline}d) models microbial communities through the generalised Lotka–Volterra (gLV) equation, utilising microbial abundance profile, along with derived data and peer-reviewed scientific literature (Fig. \ref{fig:pipeline}b,c). 
Data was derived from the abundance profile Fig. \ref{fig:pipeline}b in the form of statistics and taxonomic richness (details in Supplementary Note 5).
Text embeddings are generated by retrieving scientific literature from PubMed \cite{pubmed} relevant to the microbial taxa present in the community, and encoding using a language model (LM) \cite{minilm, bert}, as illustrated in Fig. \ref{fig:pipeline}c (details in Supplementary Note 6).

Two versions of the KIML pipeline for microbial interaction modelling have been considered in our experiments (Fig. \ref{fig:two_pipelines}): (a) multi-community generalisation pipeline and (b) mono-community pipeline.
The multi-community generalisation pipeline models interactions within multiple communities simultaneously, enabling generalisation across diverse microbial communities by capturing shared interaction dynamics and external influences. While this approach enables broader applicability (global view), it may not capture contextual knowledge unique to some communities. In contrast, the mono-community pipeline focuses on a single microbial community, enabling modelling of community-specific environments and external factors (local view). The mono-community pipeline is well-suited for studying structural changes within a community, such as the introduction of new microbial taxa, and benefits from contextual knowledge specific to the community. 

The evaluation of the modeled communities is measured by Bray-Curtis dissimilarity-based (BCD-based) accuracy \cite{imparo} and $\text{R}^2$ score \cite{baranwal2022recurrent, ruaud2024modelling} of the reconstructed abundance profile. See Section \ref{sec:methods} for more details on evaluation metrics.
For both pipelines, we evaluated our method considering three model configurations:
\begin{itemize}
    \item [(a)] A$^2$G$^2$, which does not use expert-curated knowledge and does not include gLV parameter learning
    \item [(b)] A$^2$G$^2$ combined with gLV parameter learning (APINN) guided only by microbial abundance data
    \item [(c)] A$^2$G$^2$ combined with gLV parameter learning guided by both textual information and microbial abundance data (KI-APINN)
\end{itemize}
Once the microbial community interactions are modeled, the top microbial interactions for (a), (b), and (c) are analysed and validated through MetagenomicKG \cite{metagenomickg}. 
For comparison, we use the mono-community pipeline against IMPARO \cite{imparo}, a non-neural gLV method validated on the same benchmark \cite{caporaso}, and the multi-community pipeline against the LSTM method \cite{baranwal2022recurrent} as a neural baseline.

Detailed results for each model configuration ((a)--(c)) are presented in the upcoming sections (Sections \ref{subsec:basic_vs_imparo}--\ref{subsec:unseen_gen}).
For further experiments, as described in Section \ref{subsec:other_exp}, we extended the (b) and (c) model configurations to assess the robustness and broader applicability of our approach.

\subsection{Applying A$^2$G$^2$ model to mono-community pipeline}\label{subsec:basic_vs_imparo}

\begin{figure}[h]
\centering
\includegraphics[width=1.0\textwidth]{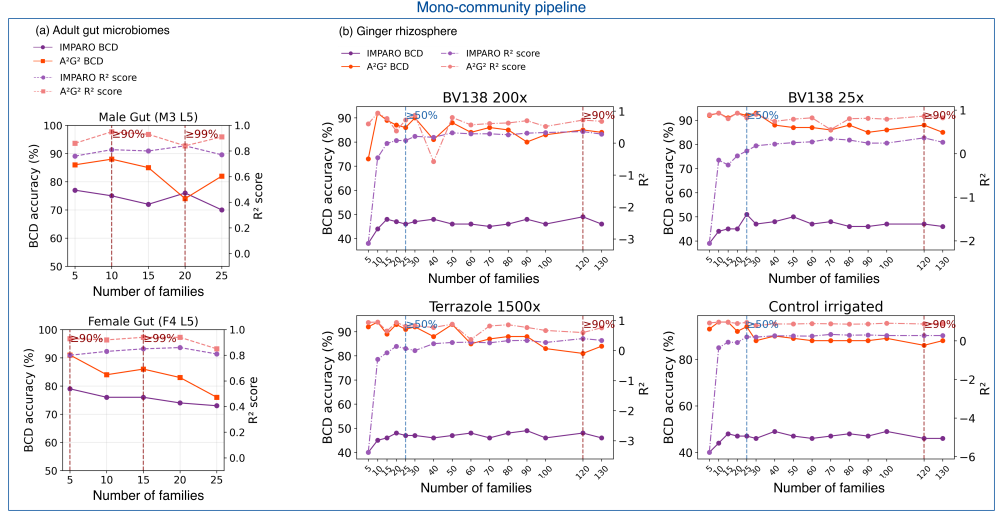}
\caption{Performance of the A$^2$G$^2$ model (mono-community pipeline) versus IMPARO across BCD accuracy and $\text{R}^2$. In each plot, the x-axis represents the number of the highest abundant families considered in each community, and the y-axes show both BCD accuracy (\%) (left) and $\text{R}^2$ score (right). For each dataset, performance is evaluated across varying community sizes in three coverage regions: $\leq$90\%, 90--99\%, and $\geq$99\% cumulative abundance for adult gut microbiomes; $\leq$50\%, 50--90\%, and $\geq$90\% cumulative abundance for ginger rhizosphere. \textbf{(a)} Performance across the adult male gut microbiome (M3) and female gut microbiome (F4) \cite{caporaso}. `L5' indicates family-level taxonomy. A$^2$G$^2$ consistently achieves higher BCD accuracy and $\text{R}^2$ than IMPARO across most regions in both datasets.
\textbf{(b)} Performance on the ginger rhizosphere \cite{soil} under different treatments. Across all treatments (BV138 200× dilution, BV138 25× dilution, Terrazole 1500× dilution, control irrigated group), A$^2$G$^2$ consistently achieves higher BCD accuracy than IMPARO. 
} 
\label{fig:imparo_vs_basic_human_ginger}
\end{figure}

For the mono-community pipeline, our A$^2$G$^2$ model was compared with IMPARO \cite{imparo} using evaluation metrics BCD accuracy and $\text{R}^2$ score (Fig. \ref{fig:imparo_vs_basic_human_ginger}). 
It is worth noting that comparisons with IMPARO are most meaningful using the mono-community pipeline results, as IMPARO is not designed for generalisation across communities \cite{copr}.

Fig. \ref{fig:imparo_vs_basic_human_ginger}a presents the performance comparison between IMPARO \cite{imparo} and A$^2$G$^2$ for adult gut microbiomes \cite{caporaso}. 

For the male gut microbiome (M3), A$^2$G$^2$ demonstrated three distinct performance patterns across coverage regions. In the $\leq$90\% region (5--10 families), A$^2$G$^2$ achieved substantial improvements over IMPARO (9--13\% for BCD accuracy, 10--14\% for $\text{R}^2$). In the 90--99\% region (10--20 families), performance degraded with increasing community size, falling slightly below IMPARO at 20 families. At $\geq$99\% coverage (25 families), A$^2$G$^2$ performance rebounded, surpassing IMPARO by 12\% for BCD and 14\% for $\text{R}^2$.

For the female gut microbiome (F4), A$^2$G$^2$ consistently outperformed IMPARO across both the 90--99\% and $\geq$99\% coverage regions, with improvements of 3--12\% for BCD and 4--13\% for $\text{R}^2$. However, the performance advantage diminished with increasing community size.

Fig. \ref{fig:imparo_vs_basic_human_ginger}b indicates the performance comparison between IMPARO \cite{imparo} and our A$^2$G$^2$ model for ginger rhizosphere for four treatments: BV138 200× dilution, BV138 25× dilution, Terrazole 1500× dilution, and the control irrigated group. 
A$^2$G$^2$ demonstrated superior BCD accuracy over IMPARO consistently across all treatments and coverage regions. For $\text{R}^2$ performance, A$^2$G$^2$ maintained superiority in nearly all cases, with one exception: BV138 200$\times$ treatment showed a slight decline below IMPARO (9\%) at 40 families. However, the magnitude of improvement varied substantially between metrics. A$^2$G$^2$ achieved substantial BCD gains of 48--53\% across all treatments at 5--10 families, while $\text{R}^2$ improvements were more modest (2--6\% range).

For detailed results for the adult gut microbiomes and ginger rhizosphere, refer to Tables S4, S5, and S6.

Mono-community pipeline results for \textit{in vitro} gut microbial communities \cite{clark2021design}, simulated gut microbial data \cite{baranwal2022recurrent}, and neonatal gut and respiratory microbiota \cite{neonatal} are provided in the Appendix. These datasets contain multiple communities and are therefore more appropriately analysed using the multi-community generalisation pipeline, which is discussed in Section \ref{subsec:glv_addition}.

A$^2$G$^2$ demonstrates robust superiority over IMPARO across diverse microbial ecosystems, achieving consistent performance gains in both gut microbiomes (upto 13\% for BCD accuracy and 14\% for $\text{R}^2$) and across all ginger rhizosphere treatments (upto 53\% for BCD accuracy and 6\% for $\text{R}^2$).

\subsection{Knowledge inclusion through gLV}\label{subsec:glv_addition}

\begin{figure}[h]
\centering
\includegraphics[width=0.95\textwidth]{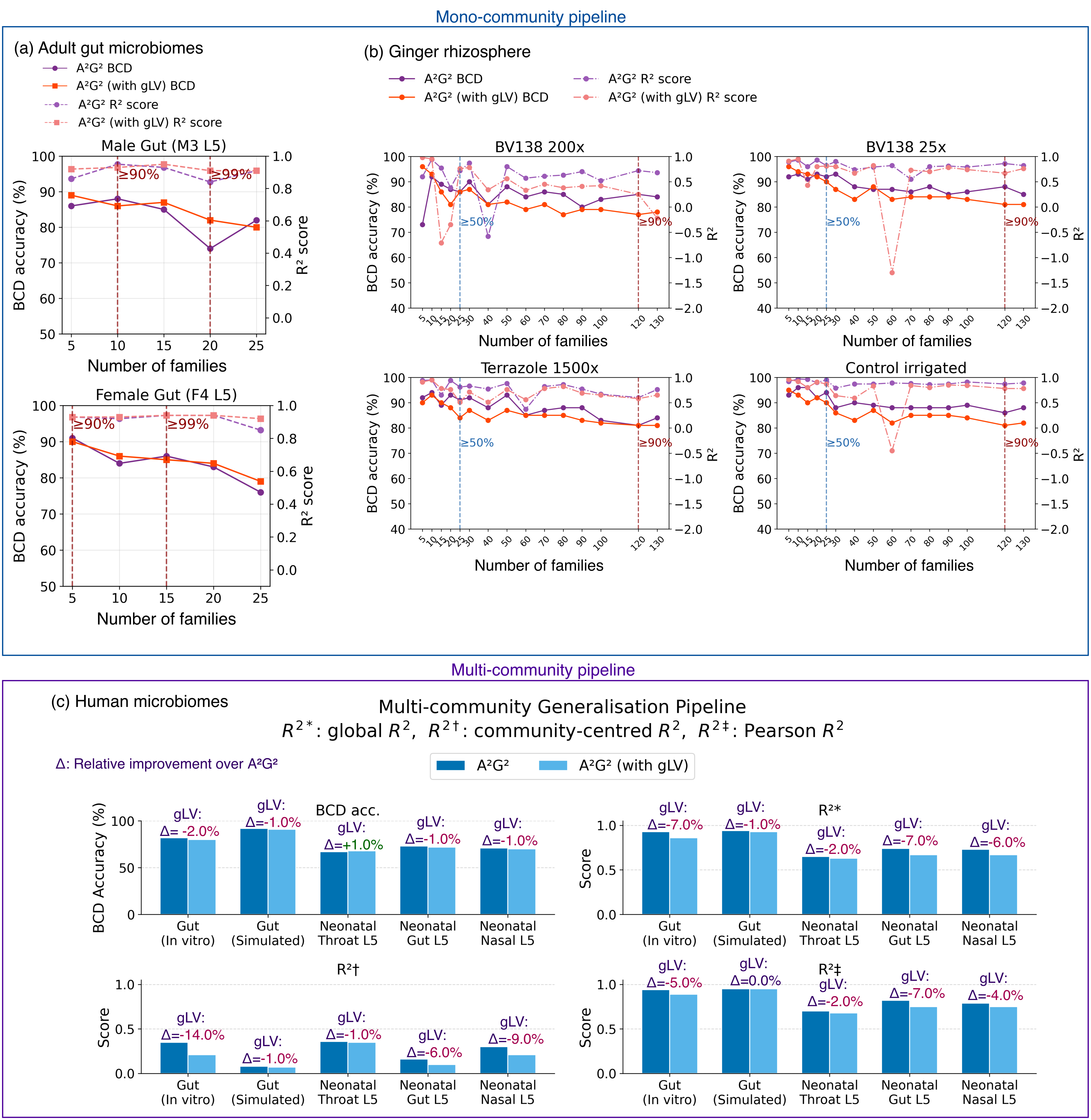}
\caption{
Performance of the A$^2$G$^2$ model versus the A$^2$G$^2$ with gLV (APINN) across BCD accuracy and $\text{R}^2$. `L5' indicates family-level taxonomy. \textbf{(a)} Performance for the adult male gut microbiome (M3) and female gut microbiome (F4) \cite{caporaso} for the mono-community pipeline. The x-axis represents the number of the highest abundant families considered in each community, and the y-axes show BCD accuracy (\%) (left) and $\text{R}^2$ score (right). For each dataset, performance is evaluated across varying community sizes in three coverage regions: $\leq$90\%, 90-99\%, and $\geq$99\% cumulative abundance.  
\textbf{(b)} Performance across the ginger rhizosphere \cite{soil} under different treatments (BV138 200× dilution, BV138 25× dilution, Terrazole 1500× dilution, control irrigated group) for the mono-community pipeline, where x-axis refers to the number of the most abundant families considered in each community. For each treatment, performance is evaluated in three coverage regions: $\leq$50\%, 50-90\%, and $\geq$90\% cumulative abundance.
\textbf{(c)} BCD accuracy and $\text{R}^2$ scores across human microbiome datasets with multiple communities for the multi-community generalisation pipeline. 
} 
\label{fig:basic_with_and_without_glv}
\end{figure}

This section compares A$^2$G$^2$ against its gLV-augmented variant, A$^2$G$^2$ (with gLV), across the two pipelines: the mono-community pipeline (Fig. \ref{fig:basic_with_and_without_glv}a,b) and the multi-community generalisation pipeline (Fig. \ref{fig:basic_with_and_without_glv}c). 

Fig. \ref{fig:basic_with_and_without_glv}a compares performance for adult gut microbiomes \cite{caporaso} in the mono-community pipeline. For the male gut microbiome (M3), gLV integration provided several benefits. First, A$^2$G$^2$ with gLV exhibited reduced performance variance across increasing families compared to base A$^2$G$^2$, demonstrating more stable behavior. Second, gLV integration recovered the BCD accuracy drop observed at 20 families in base A$^2$G$^2$ (previously noted in the A$^2$G$^2$ vs. IMPARO comparison), achieving improvements: 8\% for BCD and 7\% for $\text{R}^2$. However, A$^2$G$^2$ with gLV did not consistently outperform base A$^2$G$^2$ across all regions.

For the female gut microbiome (F4), A$^2$G$^2$ with gLV demonstrated consistent $\text{R}^2$ improvements (upto 7\% improvement). However, overall BCD accuracy was not uniformly superior across all metrics and regions.\\

Fig. \ref{fig:basic_with_and_without_glv}b presents performance comparison for the ginger rhizosphere.

\textbf{BV138 200× treatment:} gLV integration showed region-specific benefits. In the $\leq$50\% coverage region, gLV improved performance only at 5--10 families (BCD: 1--23\%, $\text{R}^2$: 1--38\%). In the 50--90\% region, base A$^2$G$^2$ generally outperformed the gLV variant except at 40 families, where gLV provided 96\% $\text{R}^2$ improvement, recovering from the performance drop where IMPARO exceeded A$^2$G$^2$ (Section~\ref{subsec:basic_vs_imparo}). In the $\geq$90\% region, base A$^2$G$^2$ consistently outperformed A$^2$G$^2$ with gLV.

\textbf{BV138 25× treatment:} gLV benefits were limited to $\leq$50\% coverage regions (1--4\% BCD, 2\% $\text{R}^2$). Base A$^2$G$^2$ was generally superior in the 50--90\% region except at 70 families (18\% $\text{R}^2$ gain). 

\textbf{Terrazole 1500× treatment:} gLV integration provided minimal benefit, with base A$^2$G$^2$ outperforming the gLV variant across nearly all regions except at 60 families in the 50--90\% region (19\% $\text{R}^2$ improvement).

\textbf{Control irrigated group:} gLV benefits were restricted to 5 families only (2\% BCD, 4\% $\text{R}^2$ improvement).

GLV integration yielded no improvement in the multi-community pipeline (Fig. \ref{fig:basic_with_and_without_glv}c), except for neonatal throat microbiome (1\% BCD accuracy gain).

In summary, gLV addition shows modest benefits for gut microbiomes versus limited and sometimes detrimental effects for ginger rhizosphere communities. This suggest that gLV dynamics alone may not universally capture the ecological principles governing different microbial ecosystems. This limitation motivates the incorporation of textual knowledge to guide gLV parameter discovery and provide ecosystem-specific context, which we explore in the following section.

\subsection{Knowledge inclusive gLV parameter discovery}\label{subsec:text_addition}

\begin{figure}[h]
\centering
\includegraphics[width=1.0\textwidth]{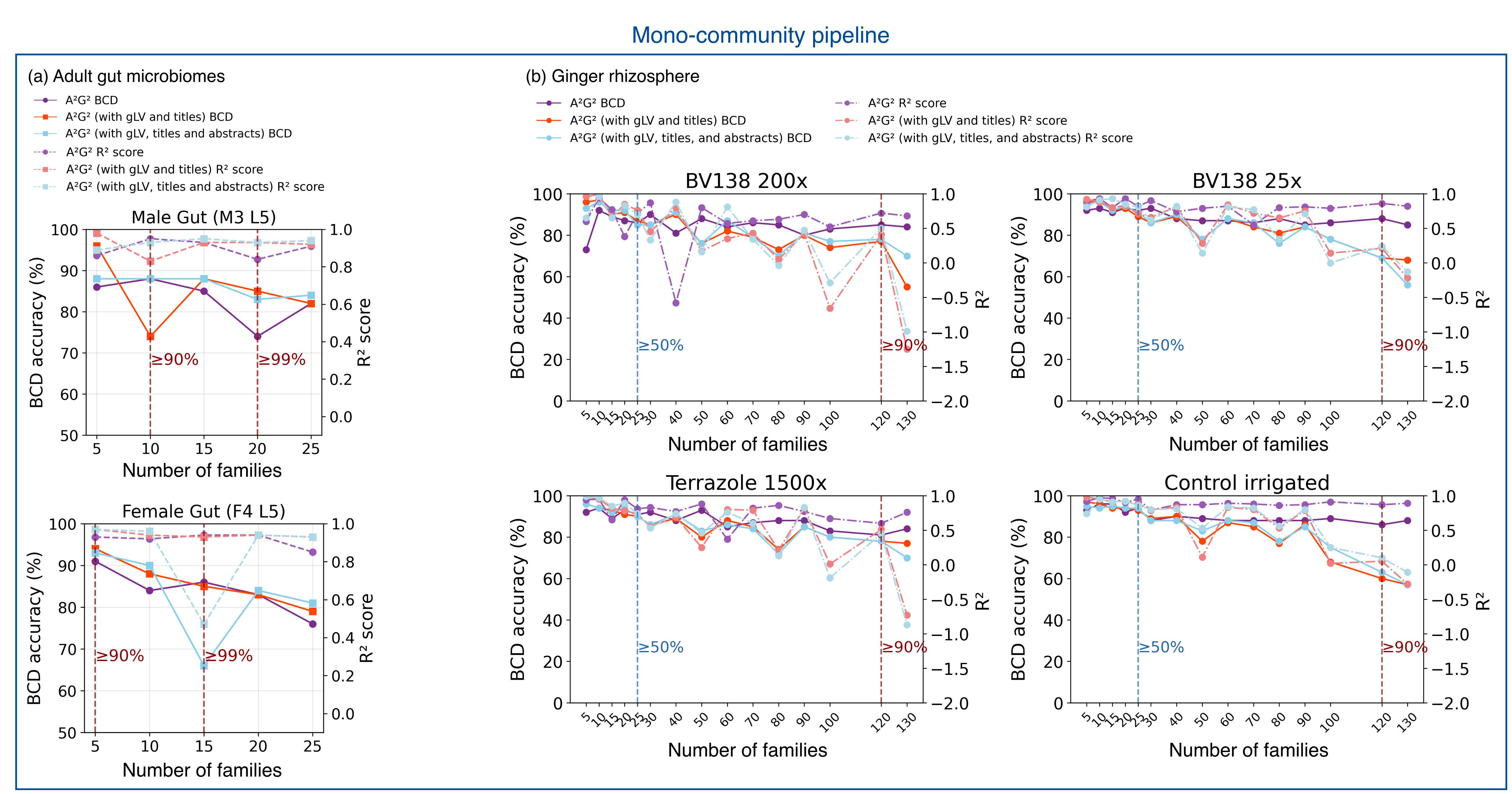}
\caption{Performance of the base A$^2$G$^2$ model versus its variants where gLV parameter learning is guided by text across BCD accuracy and $\text{R}^2$ for the mono-community pipeline. In each plot, the x-axis represents the number of the highest abundant families considered in each community, and the y-axes show both BCD accuracy (\%) (left) and $\text{R}^2$ score (right). For each dataset, performance is evaluated across varying community sizes in three coverage regions: $\leq$90\%, 90--99\%, and $\geq$99\% cumulative abundance for adult gut microbiomes; $\leq$50\%, 50--90\%, and $\geq$90\% cumulative abundance for ginger rhizosphere.
\textbf{(a)} Performance for the adult male gut microbiome (M3) and female gut microbiome (F4) \cite{caporaso}. `L5' indicates family-level taxonomy. \textbf{(b)} Performance across the ginger rhizosphere \cite{soil} under different treatments (BV138 200× dilution, BV138 25× dilution, Terrazole 1500× dilution, control irrigated group) for the mono-community pipeline.
} 
\label{fig:text_vs_basic_human_ginger}
\end{figure}

\begin{table}[h!]
\centering
\renewcommand{\arraystretch}{1.3}
\setlength{\tabcolsep}{4pt}
\begin{tabular}{|p{1.4cm}|p{1.4cm}!{\vrule width 2pt}C{1.5cm}|C{1.4cm}C{1.5cm}C{1.5cm}C{2.0cm}C{2.0cm}|}
\hline
\textbf{Dataset} 
& \textbf{Metric}
& \textbf{A$^2$G$^2$ model}
& \multicolumn{5}{c|}{\textbf{Knowledge inclusion to A$^2$G$^2$}}\\ 

 & & & \textbf{gLV (APINN)}
& \textbf{Text (titles)} 
& \textbf{Text (titles and abstracts)} 
& \textbf{gLV with text (titles) (KI-APINN)}
& \textbf{gLV with text (titles and abstracts) (KI-APINN)}\\ 
\hline

\cite{clark2021design} Gut  & BCD acc. & 82$\pm$2.6 & 80$\pm$2.8 & 82$\pm$1.7 & 81$\pm$1.8 & \textbf{83}$\pm$1.0 & 80$\pm$2.0 \\
(\textit{In vitro}) & $\text{R}^2$* & 0.93$\pm$0.02 & 0.86$\pm$0.06 & \textbf{0.96}$\pm$0.02 & 0.95$\pm$0.03 & 0.94$\pm$0.04 & \textbf{0.96}$\pm$0.01 \\ 
& $\text{R}^2$† & 0.35$\pm$0.07 & 0.21$\pm$0.11 & \textbf{0.36}$\pm$0.07 & 0.35$\pm$0.09 & 0.32$\pm$0.14 & 0.37$\pm$0.06 \\ 
& $\text{R}^2$‡ & 0.94$\pm$0.01 & 0.89$\pm$0.04 & \textbf{0.97}$\pm$0.01 & 0.96$\pm$0.01 & 0.96$\pm$0.03 & 0.96$\pm$0.01 \\ \hline

\cite{baranwal2022recurrent} Gut  & BCD acc. & \textbf{92}$\pm$1.3 & 91$\pm$2.7 & - & - & -  & -\\
(Simulated)& $\text{R}^2$* & \textbf{0.94}$\pm$0.009 & 0.93$\pm$0.02 & - & - & -  & -\\ 
& $\text{R}^2$† & \textbf{0.08}$\pm$0.06 & 0.07$\pm$0.08 & - & - & -  & -\\ 
& $\text{R}^2$‡ & \textbf{0.95}$\pm$0.006 & \textbf{0.95}$\pm$0.02 & - & - & -  & -\\ \hline

\cite{neonatal} & BCD acc. & 67$\pm$1.8 & \textbf{68}$\pm$0.5 & 67$\pm$2.2 & 65$\pm$2.4 & 67$\pm$1.9 & 67$\pm$1.9\\
Neonatal & $\text{R}^2$* & \textbf{0.65}$\pm$0.05 & 0.63$\pm$0.04 & 0.59$\pm$0.07 & 0.58$\pm$0.13 & 0.64$\pm$0.10 & 0.64$\pm$0.10 \\ 
Throat L5 & $\text{R}^2$† & 0.36$\pm$0.06 & 0.35$\pm$0.06 & 0.32$\pm$0.08 & 0.32$\pm$0.12 & 0.34$\pm$0.10 & \textbf{0.37}$\pm$0.11 \\ 
& $\text{R}^2$‡ & \textbf{0.70}$\pm$0.04 & 0.68$\pm$0.02 & 0.65$\pm$0.05 & 0.64$\pm$0.10 & \textbf{0.70}$\pm$0.07 & 0.69$\pm$0.07 \\ \hline

\cite{neonatal} & BCD acc. & 73$\pm$4.3 & 72$\pm$3.1 & 73$\pm$2.2 & \textbf{74}$\pm$2.4 & 73$\pm$2.1 & 72$\pm$1.3 \\
Neonatal & $\text{R}^2$* & 0.74$\pm$0.09 & 0.67$\pm$0.09 & 0.76$\pm$0.07 & \textbf{0.77}$\pm$0.09 & 0.72$\pm$0.07 & 0.67$\pm$0.07 \\ 
Gut L5 & $\text{R}^2$† & 0.16$\pm$0.16 & 0.10$\pm$0.16 & 0.17$\pm$0.13 & \textbf{0.21}$\pm$0.13 & 0.15$\pm$0.09 & 0.06$\pm$0.11 \\ 
& $\text{R}^2$‡ & 0.82$\pm$0.06 & 0.75$\pm$0.06 & \textbf{0.83}$\pm$0.04 & 0.82$\pm$0.07 & 0.78$\pm$0.06 & 0.75$\pm$0.05 \\ \hline

\cite{neonatal} & BCD acc. & \textbf{71}$\pm$1.5 & 70$\pm$3.5 & \textbf{71}$\pm$3.0 & 69$\pm$0.8 & 69$\pm$3.0 & 68$\pm$2.4 \\
Neonatal & $\text{R}^2$* & \textbf{0.73}$\pm$0.06 & 0.67$\pm$0.09 & 0.72$\pm$0.08 & 0.69$\pm$0.05 & 0.68$\pm$0.07 & 0.66$\pm$0.06 \\ 
Nasal L5 & $\text{R}^2$† & 0.30$\pm$0.07 & 0.21$\pm$0.14 & \textbf{0.31}$\pm$0.12 & 0.22$\pm$0.08 & 0.25$\pm$0.08 & 0.21$\pm$0.06 \\ 
& $\text{R}^2$‡ & \textbf{0.79}$\pm$0.05 & 0.75$\pm$0.05 & 0.77$\pm$0.05 & 0.77$\pm$0.04 & 0.75$\pm$0.06 & 0.74$\pm$0.05 \\ \hline

\end{tabular}
\caption{Performance comparison of human-related microbial datasets with BCD-based accuracy and $\text{R}^2$ scores ($\text{R}^2$*: global $\text{R}^2$ , $\text{R}^2$†: community-centered $\text{R}^2$, $\text{R}^2$‡: Pearson $\text{R}^2$) for the multi-community generalisation pipeline. `L5' indicates family-level taxonomy. Results represent the average performance and standard deviation across five cross-validation folds. Text inclusion was not done for the simulated gut dataset \cite{baranwal2022recurrent} (indicated by `-') due to its synthetic nature. For the \textit{in vitro} gut microbiome, text addition consistently improves performance over the base A$^2$G$^2$ (1\% - 3\%), while gLV alone provides no benefit. For neonatal microbiomes, knowledge inclusion shows mixed results: while R²† improves modestly in gut samples (5\%), most metrics either remain stable or decline slightly compared to the base A$^2$G$^2$ model. 
 }
\label{tab:methods_comparison_generalization_full}
\end{table}

We now consider the KI-APINN; A$^2$G$^2$ with gLV inclusion, where the gLV parameters are enriched with text. In this work, we use the titles and abstracts of relevant scientific papers as the source of textual knowledge, rather than full-text articles. Abstracts are designed to concisely summarise the key findings, methodology, and conclusions of a paper, and titles provide a high-level semantic signal about the subject matter. We assume that the title and abstract capture the most important information without the noise introduced by full-text content. Furthermore, incorporating full-text articles is constrained by the token limitations of current language models, which restrict the amount of text that can be processed in a single pass. We have tested the performance by gradually incorporating different expert-curated knowledge sources to observe how they contribute.

Fig. \ref{fig:text_vs_basic_human_ginger}a indicates the model performance for adult gut microbiomes \cite{caporaso}.
For the male gut microbiome (M3), text integration provided substantial benefits across most coverage regions. In the $\leq$90\% region, text integration was beneficial at 5 families (up to 10\% BCD accuracy and 12\% $\text{R}^2$ improvements) but showed diminished returns at 10 families. Both text configurations consistently outperformed base A$^2$G$^2$ in the 90--99\% and $\geq$99\% coverage regions, with BCD improvements ranging from 3--11\% and $\text{R}^2$ gains of 2--9\%. Notably, the addition of titles and abstracts exhibited the most stable performance across increasing community sizes, with reduced variance compared to other configurations.

For the female gut microbiome (F4), text integration demonstrated substantial benefits across coverage regions. In the 90--99\% coverage region, both text configurations outperformed base A$^2$G$^2$ except at 15 families. At 5--10 families, text addition achieved 2--6\% BCD improvement and 4--6\% $\text{R}^2$ improvement. In the $>$99\% coverage region, text-integrated models recovered from the performance decline at 15 families, surpassing base A$^2$G$^2$. Notably, the addition of titles exhibited the lowest variance across increasing families, demonstrating stability.

Fig. \ref{fig:text_vs_basic_human_ginger}b presents performance comparison for the ginger rhizosphere. \\
\textbf{BV138 200× treatment:} Text integration provided the strongest benefits among all treatments. In the $\leq$50\% coverage region, both text configurations substantially outperformed base A$^2$G$^2$, with BCD improvements of 1--23\% and $\text{R}^2$ gains of 2--47\%. Notably, text integration expanded the range of improvement: while A$^2$G$^2$ with gLV alone improved performance only at 5--10 families, adding text extended improvements across 5--25 families. In the 50--90\% and $\geq$90\% coverage regions, text provided minimal benefit except at 40 families.\\
\textbf{BV138 25× treatment:} Text integration showed moderate benefits concentrated in $\leq$50\% coverage region: both text configurations improved BCD accuracy modestly (1--4\%), with minimal $\text{R}^2$ gains. Notably, text integration extended the improvement range, similar to that of BV138 200×. In the 50--90\% and $\geq$90\% regions, text provided no consistent benefit except at 40 families.\\
\textbf{Terrazole 1500× treatment:} Text integration provided selective benefits in $\leq$50\% region: both text configurations outperformed base A$^2$G$^2$, with BCD improvements of 3--4\% and $\text{R}^2$ gains of 2--19\%. Notably, while gLV alone provided no improvement over base A$^2$G$^2$, adding text enabled performance gains at 5--15 families. In the 50--90\% and $\geq$90\% regions, text showed minimal benefit except at 40 and 60 families.\\
\textbf{Control irrigated group:} Text integration provided minimal benefits. In the $\leq$50\% region, improvements were restricted to 5 families only, with BCD gains of 2--4\% and $\text{R}^2$ improvement of 5\%. No consistent benefits were observed in higher coverage regions.

Refer to Tables S7 and S8 for the versions of our mono-community pipeline, which incorporates only text (without gLV) for gut microbiomes and ginger rhizosphere.\\

Table \ref{tab:methods_comparison_generalization_full} compares multi-community pipeline performance with and without expert-curated knowledge. For \textit{in vitro} gut microbiomes \cite{clark2021design}, text-guided gLV yielded modest improvements (1--3\% across metrics). For neonatal gut microbiomes \cite{neonatal}, text integration without gLV consistently improved performance across all metrics (1--5\% gains in BCD accuracy, global $\text{R}^2$, community-centered $\text{R}^2$, and Pearson $\text{R}^2$). In contrast, neonatal respiratory microbiomes (throat, nasal) showed minimal response to text integration (1\% community-centered $\text{R}^2$ gain only). Text-based experiments were not conducted for simulated gut microbiomes \cite{baranwal2022recurrent} due to their synthetic nature.

Refer Tables S9, S10, and S11 for results obtained for neonatal, \textit{in vitro} gut, and simulated gut microbiomes using the mono-community pipeline.\\

Text integration demonstrates ecosystem- and pipeline-specific benefits. In the mono-community pipeline, text consistently improved adult gut microbiome modelling. 
Text addition under the mono-community pipeline showed treatment-dependent benefits for the ginger rhizosphere (strongest for BV138 200×, weakest for control). In the multi-community pipeline, gut microbiomes again benefited the most, while respiratory microbiomes (throat, nasal) showed minimal response.

Table S12 compares our method against the LSTM approach \cite{baranwal2022recurrent} on the multi-community generalisation pipeline, where our method consistently achieves superior performance. In the LSTM approach, the interaction network is not evaluated from the model itself and is constructed by aggregating LIME explanations and then validated only by sign agreement with an external gLV model. This makes it fundamentally incomparable to methods that evaluate the quality of the interaction network using measures such as the $\text{R}^2$ score or the BCD-accuracy of the reconstructed abundance profile.
It is equally important to note that our method was not directly compared with the GNN-based approach \cite{ruaud2024modelling}, as their main objective and outputs are not fully aligned with ours and do not provide a meaningful basis for comparison. Specifically, their method does not utilise abundance data and considers only community modelling at equilibrium (a single time point), whereas our approach explicitly considers abundance dynamics and evaluates based on the abundance reconstruction.

\subsection{Generalisation to unseen taxa}\label{subsec:unseen_gen}

\begin{figure}[h]
\centering
\includegraphics[width=1.0\textwidth]{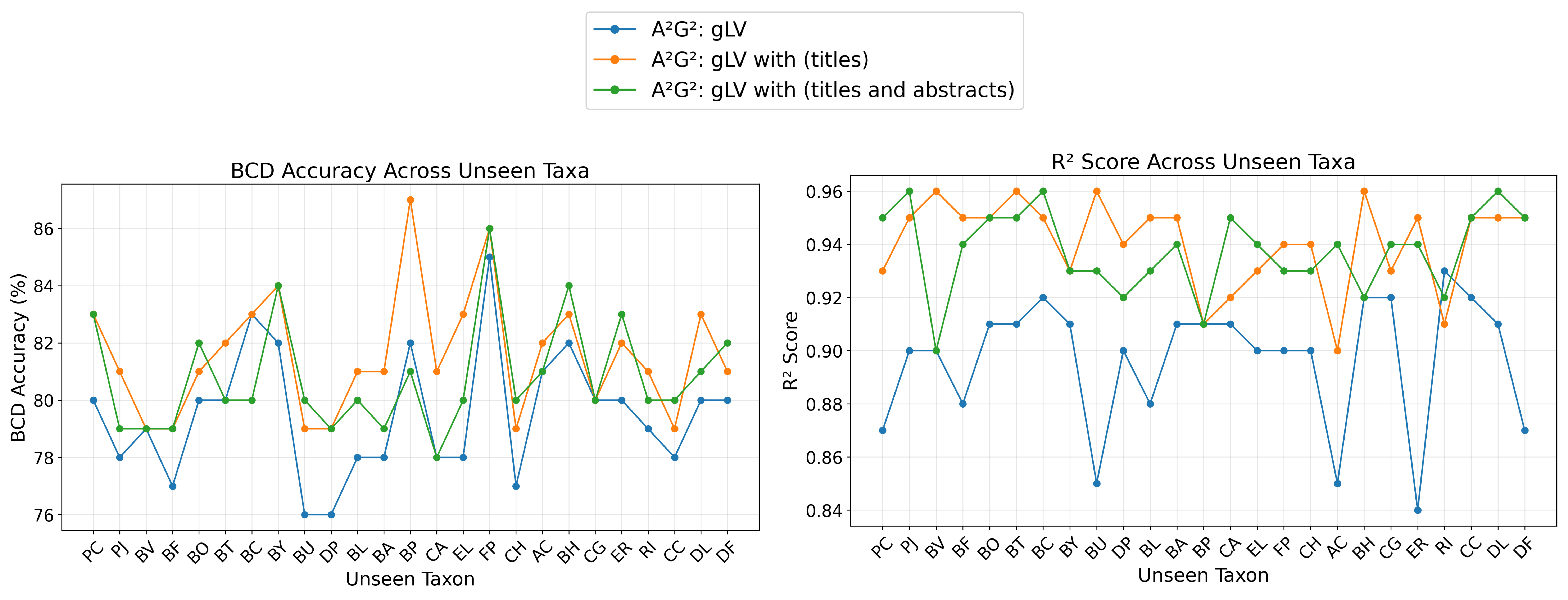}
\caption{Multi-community pipeline performance across unseen taxa for \textit{in vitro} gut microbiomes \cite{clark2021design}. BCD-based accuracy (left) and global $\text{R}^2$ score (right) for the A$^2$G$^2$ (with gLV) and its text-enriched variants using sentence embeddings. Across nearly all unseen taxa, incorporating textual information consistently improves performance over the A$^2$G$^2$ (with gLV) model. Taxonomic names corresponding to the abbreviated taxa codes are provided in Table S13.} 
\label{fig:unseen_generalization}
\end{figure}

For \textit{in vitro} gut microbial communities \cite{clark2021design}, we evaluate the performance of the multi-community pipeline under an unseen-taxa setting. This dataset comprises 25 unique species (taxonomic names corresponding to the abbreviated taxa codes are provided in Table S13), and for each experiment, we construct train–test splits such that a selected taxa is entirely held out from the training data and appears only in the test set. This enables a controlled evaluation of the model’s ability to generalise to previously unseen taxa. Model performance is then compared across the A$^2$G$^2$ (with gLV) and text-enriched variants to assess the contribution of text to generalisation across unseen taxa. The results of this experiment are shown in Fig. \ref{fig:unseen_generalization}.
Incorporating text guidance alongside abundance data for gLV learning enables A$^2$G$^2$ to consistently outperform the abundance-only variant across all 25 unseen taxa. The sole exception is \textit{R. intestinalis} (RI), where the addition of text leads to a 1–2\% performance drop in the $\text{R}^2$ score. On average, augmenting A$^2$G$^2$ (with gLV) with text yields a mean improvement of 1.3–2\% in BCD accuracy and 4.0–4.4\% in $\text{R}^2$ across all 25 unseen taxa.\\

\subsection{Network analysis of gut microbiomes}\label{subsec:gut_analysis}

\begin{figure}[h]
\centering
\includegraphics[width=1.0\textwidth]{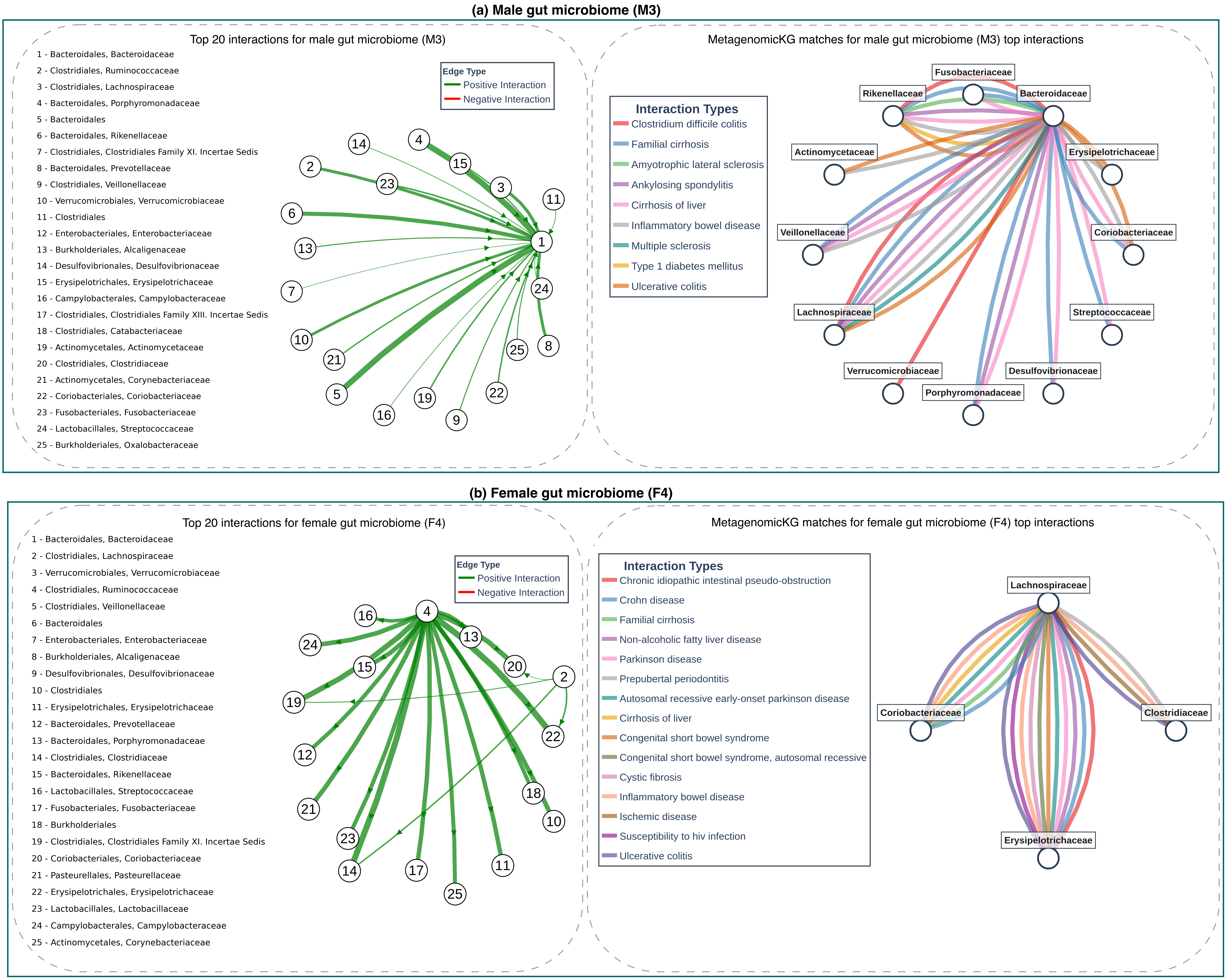}
\caption{Analysis of the inferred microbial interaction network for the male (M3) and female (F4) gut microbiomes with 25 families. For each dataset, the top 20 interactions in the modeled network (left) are shown. Nodes are labeled with the 25 most abundant family names, ranked by abundance (1 indicating the highest abundance). The thickness of the network edges denotes the interaction strengths. Green edges indicate positive interactions. Additionally, for each dataset, validation of modeled interactions against the metagenomic knowledge graph, showing disease-associated relationships between families, is indicated (right). 
\textbf{(a)} Male gut microbiome (M3): Bacteroidaceae acts as the dominant hub indicating keystone behavior. Validation of modeled interactions indicates disease-associated relationships between Bacteroidaceae and other families. 
\textbf{(b)} Female gut microbiome (F4): Ruminococcaceae acts as the dominant hub indicating keystone behavior. 
}\label{fig:m3_25_analysis_20}
\end{figure}

To understand the nature of microbial interactions in the gut, we analysed the inferred interaction networks for the human gut microbiome datasets, focusing on the male (M3) and female (F4) gut microbiomes \cite{caporaso} as representative cases (Fig. \ref{fig:m3_25_analysis_20}). Specifically, we identified the top 20 interactions within a community of 25 families ($>$99\% cumulative abundance) for both datasets. Modeled interactions were validated against MetagenomicKG \cite{metagenomickg}, a knowledge graph combining scientific literature and microbial findings (details in Supplementary Note 16).

For the male gut microbiome (M3) (Fig. \ref{fig:m3_25_analysis_20}a), Bacteroidaceae family indicates keystone behavior. All the top 20 interactions indicate positive contributions towards the Bacteroidaceae family. Multiple species in Bacteroides are known to indicate keystone behaviour \cite{shin2024bacteroides} and they have disproportionately large impacts on ecosystem dynamics \cite{culp2023cross}. Our modeled network indicates Clostridiales having positive interactions towards Bacteroidales, indicating possible cross-feeding relationships \cite{culp2023cross}. The positive interaction between Porphyromonadaceae and Bacteroidaceae, both members of the Bacteroidales order, may reflect the principle that phenotypic diversity among different families increases the fitness of the Bacteroidales community as a whole \cite{coyne2014evidence}. The positive interaction between Ruminococcaceae and Bacteroidaceae corroborates previous findings of cogrowth patterns between these families \cite{huang2023high}.
Cross-referencing with MetagenomicKG \cite{metagenomickg} revealed multiple matches from our top 20 interaction network for M3 across disease contexts. Notably, Bacteroidaceae links to multiple families across disease contexts, confirming that inferred interactions are biologically grounded and consistent with known disease-associated patterns.

For the female gut microbiome (F4) (Fig. \ref{fig:m3_25_analysis_20}b), Ruminococcaceae acts as a hub and interacts positively with other families. The positive interaction between Ruminococcaceae and Coriobacteriaceae is consistent with observed cogrowth patterns between them \cite{huang2023high}. Cross-referencing with MetagenomicKG revealed interactions between Lachnospiraceae and multiple families.

For an in-depth analysis, the top 50 interactions for both gut microbiome datasets are presented in Fig. S1. The distribution of pairwise interaction strengths between all families in the inferred networks, attention scores learned for input modalities, and growth rates for each family are provided in Fig. S2.

\subsection{Network analysis of ginger rhizosphere communities}\label{subsec:ginger_analysis}

\begin{figure}[htbp]
\centering
\includegraphics[width=1.0\textwidth]{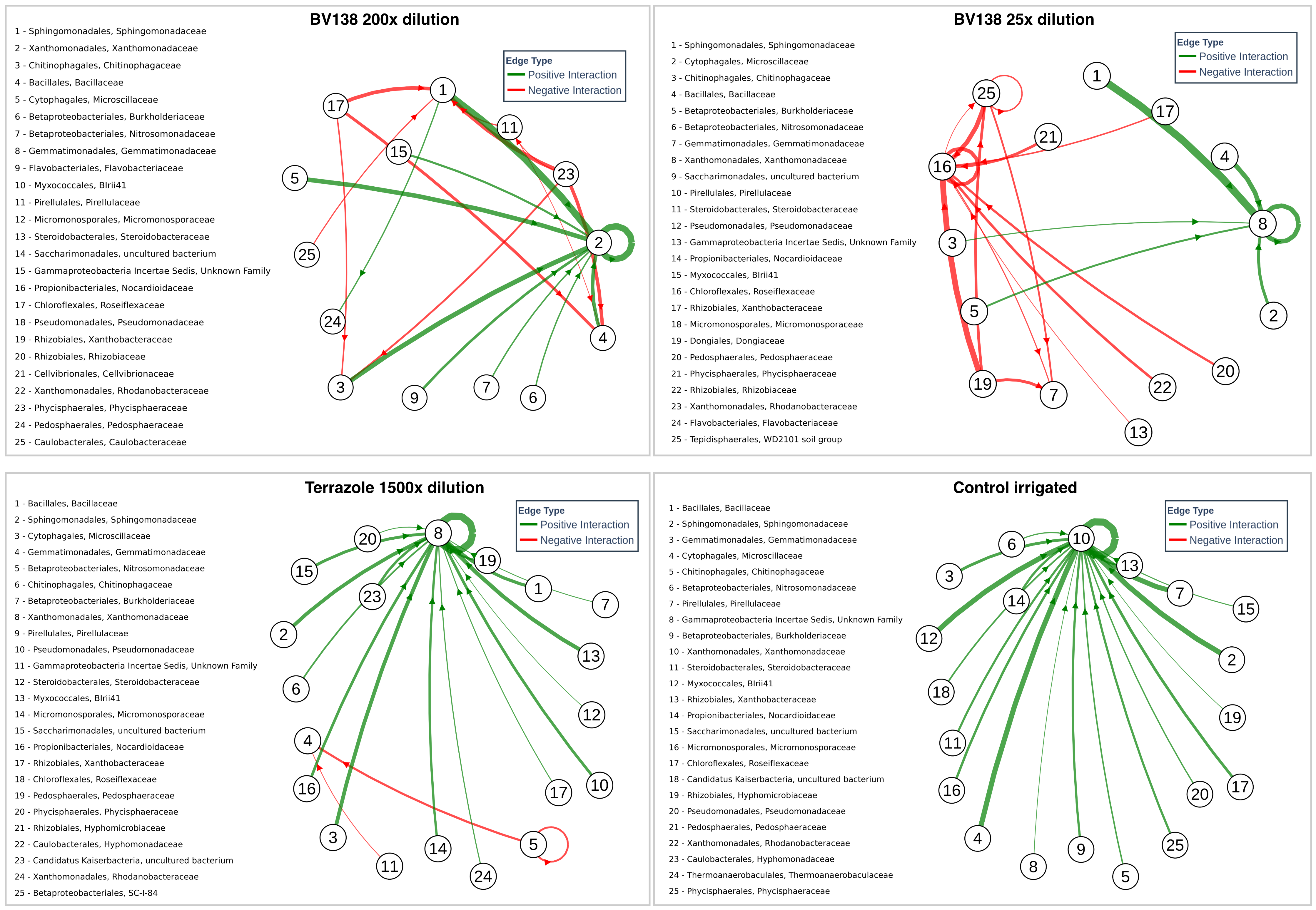}
\caption{We conducted interaction network analysis for the ginger rhizosphere community across four treatments: BV138 200× dilution, BV138 25× dilution, Terrazole 1500× dilution, and control irrigated. For each treatment, we identified and characterised the top 20 strongest microbial interactions within communities comprising 25 microbial families (families are ranked based on their abundance levels; 1 being the most abundant in that community). In the interaction networks, red edges denote negative interactions while green edges denote positive interactions. The thickness of the network edges denotes the interaction strengths.}
\label{fig:ginger_analysis_20}
\end{figure}

Fig. \ref{fig:ginger_analysis_20} indicates the inferred interaction networks for the four treatments in the ginger rhizosphere \cite{soil}. We identified the top 20 interactions within a community of 25 families (approximately 50\% cumulative abundance coverage) for each treatment. These interactions were validated through direct literature search rather than MetagenomicKG \cite{metagenomickg}, which is specific to human microbiomes.

BV138 200× dilution:
Xanthomonadaceae exhibited keystone/hub behavior, receiving positive support from multiple families: Sphingomonadaceae, Chitinophagaceae, Nitrosomonadaceae, and Bacillaceae. Sphingomonadaceae displayed secondary hub behavior alongside Xanthomonadaceae. The keystone roles of both families are supported by previous work \cite{cho2024ecological}, which identified seven distinct microbial families, including Sphingomonadaceae and Xanthomonadaceae, as an `initial community' participating in soil nitrogen metabolism and regulating root rot disease through nutritional competition during ginseng monoculture. Additionally, Xanthomonadaceae has been identified as a keystone taxon in wheat rhizosphere communities \cite{acuna2023diversity}. Roseiflexaceae emerged as an important family, exhibiting negative interactions with Sphingomonadaceae, Chitinophagaceae, and Bacillaceae.

BV138 25× dilution:
Xanthomonadaceae maintained keystone/hub behavior, receiving support from other families. Roseiflexaceae also exhibited keystone/hub behavior, characterised by negative interactions from multiple families.

Terrazole 1500× dilution:
Xanthomonadaceae continued to display keystone/hub behavior, receiving support from other families. Gemmatimonadaceae emerged as an important node, receiving negative interactions from Nitrosomonadaceae and Gammaproteobacteria Incertae Sedis.  

Control irrigated:
Xanthomonadaceae maintained keystone/hub behavior, receiving support from multiple families. This cooperative network is consistent with previous findings \cite{wang2026microbial}, which demonstrated that Gemmatimonadaceae and Xanthomonadaceae were significantly enriched in the rhizosphere and emerged as dominant microbial taxa regulating the microecosystem, with their enrichment associated with reduced pathogenic fungal populations. 

For an in-depth analysis, the top 50 interactions for each treatment are presented in Fig. S3. The distribution of pairwise interaction strengths between all families, attention scores learned for input modalities, and growth rates for each family are provided in Fig. S4.

\subsection{Taxonomic scaling, text quality enhancement, and knowledge graph integration}\label{subsec:other_exp}

Experiments were conducted on male (M3) and female (F4) gut microbiomes \cite{caporaso} examining: (1) performance across taxonomic levels, (2) text quality enhancement via Q1-ranked journal filtering, (3) biomedical domain specificity through PubMedBERT \cite{pubmed} embeddings, and (4) structured knowledge integration using MetagenomicKG \cite{metagenomickg} as \textbf{network edge embeddings}. 
For (1), the robustness of our method was evaluated across different taxonomic levels: genus, family, order, class, and phylum.
For (2), we compared the usage of text obtained from only Q1 journals against the text used in the previous sections. 
For (3), we compare text embeddings obtained using ``PubMedBert" \cite{bert} (domain-specific LM) against the previously used text embeddings from ``all-MiniLM-L6-v2" \cite{minilm} (general LM). 
For (4), we assessed the incremental benefit of adding Knowledge Graph (KG) embeddings, in the form of edge embeddings, to the three configurations: gLV alone, gLV with titles, and gLV with titles and abstracts.\\

Fig. S5 indicates evaluation across different taxonomic levels. Incorporating text information to guide gLV parameter discovery either improved or matched performance across all taxonomic levels in both datasets.

Restricting text embeddings to Q1-ranked journals (Table S14) modestly improved male gut performance (up to 13\% BCD accuracy gain at 10 families) but degraded female gut performance, suggesting that high-impact filtering may exclude relevant microbial literature.
PubMedBERT showed mixed results (Table S15): modest improvements for male gut at smaller communities but severe degradation at 25 families when using both titles and abstracts. Female gut performance was more stable but did not consistently improve over baseline.

Integration of knowledge graph edge embeddings (Table S16) demonstrated consistent performance improvements across both gut datasets. KG embeddings alone yielded stable gains of 2--3\% BCD accuracy improvement for adult gut microbiomes.
Combined with textual information, knowledge graphs exhibited gains up to 19\% for BCD accuracy and 47\% for $\text{R}^2$ for adult gut microbiomes.

\section{Discussion}\label{sec:discussion}

We introduced Knowledge Inclusive Adaptive PINN (KI-APINN), a framework that systematically incorporates contextual knowledge into physics-informed neural networks. We applied this framework to microbial interaction modelling using our A$^2$G$^2$ as the underlying neural architecture. In this application, KI-APINN learns gLV parameters by jointly leveraging both microbial abundance data and metagenomic literature.

Interpretability of the model is achieved through several ways: (a) the attention mechanism that describes the importance given to different inputs to the model for each taxon in the community, and (b) modelling the gLV equation for each community while learning growth and interaction parameters of taxa of the microbial network. The ability to easily observe these outputs after the modelling process provides user-friendliness to the method.

It is important to note that the growth rates and pairwise interactions learned by the model should be interpreted as comparative rather than absolute values. Their biological significance lies primarily in their sign, indicating whether interactions are competitive or supportive, and their relative magnitudes, which reflect the strength of one interaction compared to another. No constraints are imposed on the parameters during learning beyond the structural form of the gLV equation itself, as the true parameter ranges are unknown for the communities considered. 

Our method, even without knowledge inclusion, demonstrated robust superiority over IMPARO across gut and rhizosphere microbial ecosystems, with BCD accuracy gains up to 53\%. This consistent performance across ecosystem types, treatment conditions, and host subjects indicates that our A$^2$G$^2$'s architecture effectively captures fundamental microbial dynamics across diverse ecological contexts.

Integration of generalised Lotka-Volterra (gLV) dynamics demonstrated ecosystem-dependent effects. 
For mono-community evaluation, gLV improved gut microbiome performance \cite{caporaso} but only benefited ginger rhizosphere \cite{soil} at low coverage ($\leq$50\% cumulative abundance); base A$^2$G$^2$ outperformed at higher coverage.
 
These contrasting patterns suggest that gLV dynamics are often effective for gut microbiome interactions but may be insufficient for rhizosphere microbial communities. This may be attributed to soil microbiomes having more evenly distributed abundance profiles across taxa (higher community evenness), which increases the complexity of the modelling problem compared to gut communities, where a few dominant taxa often account for most of the biomass \cite{gonze2018microbial}. 

Using text to guide the gLV parameter learning process demonstrated clear ecosystem specificity, with consistent benefits for gut microbiomes but variable effects elsewhere. Text integration requirements varied by community: the male microbiome \cite{caporaso} improved with both titles and abstracts, whereas the female microbiome \cite{caporaso} benefited from titles alone, demonstrating that optimal information sources depend on community composition. 
For \textit{in vitro} \cite{clark2021design} gut communities, text integration enabled robust generalisation to unseen taxa, yielding improvements of up to 4.4\% when each of 25 species was individually excluded from training and tested, demonstrating that biological knowledge encoded in text complements abundance-based learning.
Ginger rhizosphere communities \cite{soil} showed treatment-dependent text benefits, strongest for BV138 200× dilution (upto 23\% for BCD accuracy and 47\% for $\text{R}^2$) and weakest for control conditions.
Neonatal respiratory microbiomes (throat, nasal) \cite{neonatal} showed minimal response to text integration despite substantial neonatal gut microbiome benefits, suggesting that documented knowledge about respiratory microbiome interactions may be less applicable to modelling than for gut communities. Neonatal microbiomes present inherent modelling challenges due to their developmental dynamics and temporal instability \cite{neu2025neonatal}, which may contribute to the limited benefits observed from text integration across neonatal body sites.

Network analysis revealed biologically meaningful patterns consistent with established principles. In the male gut microbiome \cite{caporaso}, Bacteroidaceae exhibited keystone behavior consistent with documented roles of Bacteroides species \cite{shin2024bacteroides}. The positive interaction between Porphyromonadaceae and Bacteroidaceae, both members of the Bacteroidales order, may reflect the principle that phenotypic diversity among different families increases the fitness of the Bacteroidales community as a whole \cite{coyne2014evidence}. 
In the female gut microbiome \cite{caporaso}, Ruminococcaceae emerged as a central hub with positive interactions with Coriobacteriaceae, matching documented cogrowth patterns \cite{huang2023high}. For ginger rhizosphere communities, Xanthomonadaceae consistently displayed keystone behavior across all four treatments, with Sphingomonadaceae and Chitinophagaceae providing stable support in every treatment. This inferred network aligns with documented roles of Chitinophagaceae in producing antifungal enzymes \cite{xu2025functional}. Roseiflexaceae emerged as an important node in both BV138 200× and 25× dilution treatments, suggesting dilution-specific ecological relevance. While these core interactions remained stable, treatment-specific secondary interactions emerged depending on environmental conditions: Nitrosomonadaceae supported Xanthomonadaceae only in BV138 200× dilution and control treatments, but was absent from BV138 25× dilution and Terrazole 1500× treatments. 

Performance remained strong across taxonomic levels (genus to phylum), with consistent text integration benefits.
Q1 journal filtering and using PubMedBERT embeddings \cite{bert} had inconsistent benefits. Knowledge graph integration as a knowledge source for gLV parameter learning provided the most consistent benefits. These results suggest that text source and representation choices must account for community composition, with structured biological knowledge offering more reliable improvements than unstructured text alone.

For improving the microbial interaction modelling process, future work will include experimenting on the structural modification of the classical gLV equation to further capture microbial interactions and expanding the KI-APINN to other applicable areas.

\section{Methods}\label{sec:methods}

\subsection{Preliminaries and Background Concepts}

We first introduce the notation and background concepts.\\

\textbf{Microbial community abundance data:}
Let $\mathcal{V} = \{1, \dots, n\}$ denote the set of $n$ microbial taxa in a community. For each taxon $v \in \mathcal{V}$, we observe its relative abundance over $T$ discrete time points. The full abundance matrix is $X \in \mathbb{R}^{n \times T}$, where $X_{v,t}$ denotes the abundance of taxon $v$ at time $t$.\\ 

\textbf{Auxiliary knowledge sources:}
Beyond abundance data, we employ three auxiliary knowledge sources for each taxon.
\begin{itemize}
  \item \emph{Taxonomic richness.} The matrix $R \in \mathbb{R}^{n \times d_R}$ collects per-taxon richness statistics, where $R_v \in \mathbb{R}^{d_R}$ is the richness vector of taxon $v$ computed at a prescribed taxonomic resolution.
  \item \emph{Text embeddings.} The matrix $E \in \mathbb{R}^{n \times d_E}$ collects per-taxon text representations, where $E_v \in \mathbb{R}^{d_E}$ is a fixed embedding of peer-reviewed metagenomics literature relevant to taxon $v$, produced by a pretrained language model.
  \item \emph{Interaction graph.} The adjacency matrix $A \in \{0,1\}^{n \times n}$ encodes prior knowledge of microbial interactions, where $A_{uv} = 1$ indicates a known or inferred interaction between taxa $u$ and $v$. In this work, we consider a fully-connected graph with self-loops.
\end{itemize}

\textbf{Generalised Lotka-Volterra model:}
The generalised Lotka-Volterra (gLV) equation governs the temporal dynamics of taxon $v$ as
\begin{equation}\label{eq:glv}
  \frac{dX_{v,t}}{dt}
  \;=\; X_{v,t}\!\left(r_v + \sum_{u \in \mathcal{V}} B_{vu}\, X_{u,t}\right),
\end{equation}
where $r_v \in \mathbb{R}$ is the intrinsic growth rate of taxon $v$ and $B_{vu} \in \mathbb{R}$ quantifies the effect of taxon $u$ on taxon $v$. The parameter vector $\bm{\theta} = (r, B)$, with $r \in \mathbb{R}^n$ and $B \in \mathbb{R}^{n \times n}$, is the primary target of inference.\\

\textbf{Physics-Informed Neural Networks:}
Let $\mathcal{F}[\,\cdot\,\mathbin{;}\bm{\theta}]$ denote a differential operator parameterised by $\bm{\theta}$. A Physics-Informed Neural Network (PINN) introduces a neural surrogate $u_\phi \mathbin{:} \mathcal{X} \to \mathcal{Y}$
with learnable weights $\phi \in \mathbb{R}^{d_\phi}$,
approximating the true solution $u$ of
$\mathcal{F}[u\mathbin{;}\bm{\theta}] = 0$, and is trained by minimising a composite objective that penalises both data fitting error and violation of the physics constraint.

\subsection{Mathematical formulation}

\subsubsection{Standard PINN for parameter discovery}

Given a training dataset $\mathcal{D} = \{(x_i, y_i)\}_{i=1}^{N_d}$ of input-output pairs and a set of $N_p$ collocation points sampled from the problem domain $\{x_j\}_{j=1}^{N_p}$ at which the physics constraint is enforced, the standard parameter-discovery PINN objective is
\begin{equation}\label{eq:pinn}
  \mathcal{L}_{\mathrm{PINN}}(\phi, \bm{\theta})
  \;=\;
  \underbrace{
    \frac{1}{N_d}\sum_{i=1}^{N_d}
    \bigl\|u_\phi(x_i) - y_i\bigr\|^2
  }_{\mathcal{L}_{\mathrm{Data}}(\phi)}
  \;+\;
  \lambda\,
  \underbrace{
    \frac{1}{N_p}\sum_{j=1}^{N_p}
    \bigl\|\mathcal{F}[u_\phi\mathbin{;}\bm{\theta}](x_j)\bigr\|^2
  }_{\mathcal{L}_{\mathrm{Physics}}(\phi, \bm{\theta})},
\end{equation}
where $\lambda > 0$ is a trade-off hyperparameter and $\bm{\theta}$ is estimated jointly with the surrogate weights $\phi$ from the experimental data $\mathcal{D}$ alone.

\subsubsection{KI-APINN: Knowledge-Inclusive Parameter Discovery}

\textbf{Knowledge encoders:}
Let $\mathcal{K} = \{K_1, \dots, K_M\}$ be a collection of $M$
auxiliary knowledge sources. For each $m \in \{1, \dots, M\}$,
an encoder $g_m$ maps the $m$-th source to a fixed-dimensional
representation $g_m(K_m) \in \mathbb{R}^{d_m}$.\\

\textbf{Knowledge-inclusive representation:}
The experimental input $x \in \mathbb{R}^{d_x}$, with $d_x = nT$, represents the flattened abundance matrix for one community observation drawn from $X$. It is fused with the $M$ knowledge embeddings by a learnable fusion operator
$\Phi \mathbin{:} \mathbb{R}^{d_x} \times \mathbb{R}^{d_1} \times \cdots
\times \mathbb{R}^{d_M} \to \mathbb{R}^{d_z}$
to produce the knowledge-inclusive (KI) representation
\begin{equation}\label{eq:fusion}
  z \;=\; \Phi\!\left(x,\, g_1(K_1), \dots, g_M(K_M)\right),
  \qquad z \in \mathbb{R}^{d_z},
\end{equation}
where $d_z$ is the dimension of the KI representation.\\

\textbf{Parameter network:}
We parameterise $\bm{\theta}$ as the output of a learned map
$\bm{\theta}_\psi \mathbin{:} \mathbb{R}^{d_z} \to \Theta$,
where $\psi \in \mathbb{R}^{d_\psi}$ denotes the weights of the
parameter network, $d_\psi$ is their total count, and $\Theta$ is the admissible parameter space. The physics parameters are then given by $\bm{\theta} = \bm{\theta}_\psi(z)$.\\

\textbf{KI-APINN objective:}
The full training objective is
\begin{equation}\label{eq:ki-apinn}
  \mathcal{L}_{\mathrm{KI\text{-}APINN}}(\phi, \psi)
  \;=\;
  \mathcal{L}_{\mathrm{Data}}(\phi)
  \;+\;
  \lambda\,\mathcal{L}_{\mathrm{Physics}}\!\left(\phi,\,
    \bm{\theta}_\psi(z)\right)
  \;+\;
  \mu\,\Omega(\psi),
\end{equation}
where $\mu \geq 0$ is a balancing coefficient and
$\Omega(\psi) = \|\psi\|_2^2$ is a regularisation term applied on framework learning.\\

\textbf{Reduction to standard PINN:}
When $M = 0$ (no auxiliary knowledge) and $\bm{\theta}_\psi \equiv \bm{\theta}$ (i.e., $\psi$ is a direct free-parameter vector), equation~\eqref{eq:ki-apinn} reduces exactly to equation~\eqref{eq:pinn}. The framework is model-agnostic in that the surrogate $u_\phi$, fusion operator $\Phi$, and parameter map $\bm{\theta}_\psi$ are all interchangeable design choices. The novelty of our work lies in conditioning physics parameters on knowledge-inclusive
representations via the map $\bm{\theta}_\psi(z)$.

\subsubsection{A\textsuperscript{2}G\textsuperscript{2}: concrete instantiation for metagenomics}

$\mathrm{A}^2\mathrm{G}^2$ is the concrete realisation of
KI-APINN for microbial interaction modelling.
The architecture is the composition
\begin{equation}\label{eq:composition}
  f \;=\; h \circ s \circ a,
\end{equation}
where the feature encoder
$a \mathbin{:} \mathbb{R}^{n \times T} \times
\mathbb{R}^{n \times d_R} \times \mathbb{R}^{n \times d_E} \to
\mathbb{R}^{n \times d_h}$
maps all input modalities to initial node embeddings,
the structural encoder
$s \mathbin{:} \mathbb{R}^{n \times d_h} \times \{0,1\}^{n \times n} \to \mathbb{R}^{n \times d_h}$
refines them via graph message passing, and the reconstruction head $h \mathbin{:} \mathbb{R}^{n \times d_h} \to \mathbb{R}^n \times \mathbb{R}^{n \times n}$
decodes the gLV parameters. The three modules are described below.

\begin{figure}[h]
\centering
\includegraphics[width=0.85\textwidth]{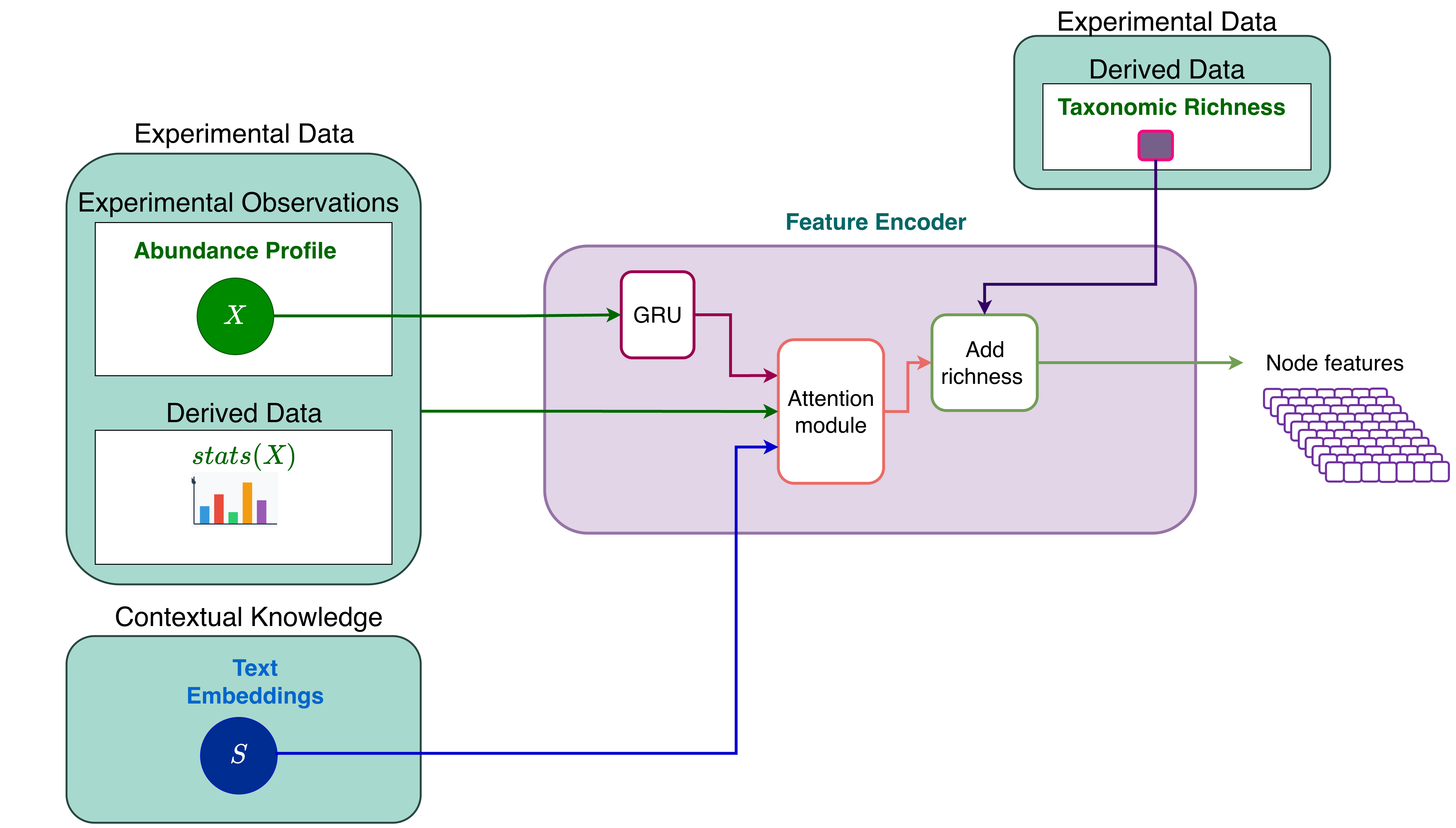}
\caption{Feature Encoder architecture of the A$^2$G$^2$ model takes different data and knowledge types as input: (1) experimental observations (abundance profile), (2) derived data (statistics and taxonomic richness), and (3) contextual knowledge (text embeddings from scientific literature). An attention module is used to fuse these modal inputs. The output is given as of node-level features for downstream structural encoding.
}
\label{fig:encoder}
\end{figure}

\textbf{(i) Attention-based feature encoder $a$.}

Fig. \ref{fig:encoder} denotes the architecture of the feature encoder.
Inspired by the work done in the LSTM-based microbial interaction modelling \cite{baranwal2022recurrent}, in order to capture the temporal dynamics in the abundance profile, we include a Gated Recurrent Unit (GRU) within the feature encoder.

Let $\mathrm{GRU}_\phi \mathbin{:} \mathbb{R}^T \to \mathbb{R}^{d_h}$
denote a GRU with hidden dimension $d_h$,
whose parameters form part of $\phi$.
For each taxon $v$, the temporal abundance sequence
$X_{v,1:T} \in \mathbb{R}^T$ is processed by the GRU to obtain a temporal summary vector
\begin{equation}\label{eq:gru}
  \tilde{x}_v \;=\; \mathrm{GRU}_\phi(X_{v,1:T}),
  \qquad \tilde{x}_v \in \mathbb{R}^{d_h}.
\end{equation}
Let $\mathcal{M} = \{x, s, e\}$ index the three attention modalities:
GRU-encoded abundance ($x$), projected statistics ($s$), and projected text embeddings ($e$). Their representations are
\begin{equation}
  \tilde{m}_v^{(x)} = \tilde{x}_v, \qquad
  \tilde{m}_v^{(s)} = W_s S_v, \qquad
  \tilde{m}_v^{(e)} = W_e E_v,
\end{equation}
where $S_v \in \mathbb{R}^{d_S}$ is the statistics vector of taxon $v$,
$W_s \in \mathbb{R}^{d_h \times d_S}$ and
$W_e \in \mathbb{R}^{d_h \times d_E}$ are learnable projection matrices.
The attention weight for modality $m \in \mathcal{M}$ is
\begin{equation}\label{eq:attn}
  \alpha_v^{(m)}
  \;=\;
  \frac{
    \exp\!\left(q^\top \tilde{m}_v^{(m)}\right)
  }{
    \displaystyle\sum_{m' \in \mathcal{M}}
    \exp\!\left(q^\top \tilde{m}_v^{(m')}\right)
  },
\end{equation}
where $q \in \mathbb{R}^{d_h}$ is a learned global query vector. The attention-fused embedding is
\begin{equation}\label{eq:fused}
  \bar{H}_v
  \;=\;
  \sum_{m \in \mathcal{M}} \alpha_v^{(m)}\, \tilde{m}_v^{(m)},
  \qquad \bar{H}_v \in \mathbb{R}^{d_h}.
\end{equation}
Taxonomic richness bypasses the attention module and is incorporated by a residual addition after a linear projection,
\begin{equation}\label{eq:richness}
  H_v^{(0)}
  \;=\;
  \bar{H}_v + W_r R_v,
  \qquad H_v^{(0)} \in \mathbb{R}^{d_h},
\end{equation}
where $W_r \in \mathbb{R}^{d_h \times d_R}$ is a learnable projection matrix. Stacking over all taxa yields
$H^{(0)} \in \mathbb{R}^{n \times d_h}$,
the initial node feature matrix passed to the structural encoder.

Ablations for the feature encoder inputs and components are presented in Table S17.

\textbf{(ii) AnchorGNN structural encoder $s$.} 

To capture the structural interactions between microbes, we employ a task-adaptive anchor-based GNN architecture. To reduce the quadratic cost of full-graph message passing, a learned scoring function assigns an anchor score
\begin{equation}\label{eq:anchor-score}
  \pi_v \;=\; \sigma\!\left(w^\top H_v^{(0)}\right),
  \qquad \pi_v \in (0,1),
\end{equation}
to each taxon, where $w \in \mathbb{R}^{d_h}$ is a learnable weight vector and $\sigma$ is the sigmoid function. The anchor set $\mathcal{A} \subset \mathcal{V}$ with $|\mathcal{A}| = k \ll n$ is formed by retaining the $k$ taxa with the highest scores.

The original adjacency $A$, which is fully-connected, is sparsified to retain only edges incident to at least one anchor node,
\begin{equation}\label{eq:sparse-adj}
  \tilde{A}_{uv}
  \;=\;
  A_{uv} \cdot
  \mathbb{1}\!\left[u \in \mathcal{A} \;\text{or}\; v \in \mathcal{A}\right].
\end{equation}
Let $\hat{A} = \tilde{A} + I_n$ denote the self-loop-augmented
adjacency, where $I_n \in \mathbb{R}^{n \times n}$ is the identity matrix, and let $\hat{D} \in \mathbb{R}^{n \times n}$ be the corresponding diagonal degree matrix with entries
$\hat{D}_{vv} = \sum_{u} \hat{A}_{vu}$.
Message passing is applied over $L \in \mathbb{Z}_{>0}$ layers,
\begin{equation}\label{eq:mp}
  H^{(\ell+1)}
  \;=\;
  \sigma_{\mathrm{act}}\!\left(
    \hat{D}^{-1/2}\hat{A}\hat{D}^{-1/2}\,
    H^{(\ell)}\, W^{(\ell)}
  \right),
  \qquad \ell = 0, 1, \dots, L-1,
\end{equation}
where $W^{(\ell)} \in \mathbb{R}^{d_h \times d_h}$ is the layer-wise learnable weight matrix and $\sigma_{\mathrm{act}}$ is a nonlinear activation function. The per-layer time complexity reduces from $\mathcal{O}(n^2 d_h)$ for a dense graph to $\mathcal{O}(n k d_h)$ for the sparsified graph.

Ablation studies were conducted to evaluate GNN architecture choices. As shown in Table S18, AnchorGNN consistently outperforms both GAT \cite{velivckovic2017graph} and GraphSAGE \cite{hamilton2017inductive}, while the no-GNN (MLP) baseline demonstrates that graph structure learning provides substantial benefit to the framework, especially in settings where text is not included as input.

\textbf{(iii) gLV reconstruction head $h$.}

The final node embeddings $H^{(L)} \in \mathbb{R}^{n \times d_h}$, with $H_v^{(L)}$ denoting the row corresponding to taxon $v$, constitute the instantiation of the KI representation $z$ in equation~\eqref{eq:fusion}, that is, $z = H^{(L)}$ for $\mathrm{A}^2\mathrm{G}^2$.
Two multilayer perceptrons,
$\mathrm{MLP}_r \mathbin{:} \mathbb{R}^{d_h} \to \mathbb{R}$
and
$\mathrm{MLP}_B \mathbin{:} \mathbb{R}^{2d_h} \to \mathbb{R}$,
decode per-taxon growth rates and pairwise interaction strengths,
\begin{equation}\label{eq:decode}
  \hat{r}_v
  \;=\; \mathrm{MLP}_r\!\left(H_v^{(L)}\right),
  \qquad
  \hat{B}_{uv}
  \;=\; \mathrm{MLP}_B\!\left(\left[H_u^{(L)} \,\|\, H_v^{(L)}\right]\right),
\end{equation}
where $[\,\cdot \,\|\, \cdot\,]$ denotes vector concatenation,
$\hat{r}_v \in \mathbb{R}$ and $\hat{B}_{uv} \in \mathbb{R}$.

The physics residual is computed as
\begin{equation}\label{eq:phys-loss}
  \mathcal{L}_{\mathrm{Physics}}(\hat{r}, \hat{B})
  \;=\;
  \frac{1}{nT}
  \sum_{v \in \mathcal{V}}\,\sum_{t=1}^{T}
  \left(
    \dot{X}_{v,t}
    \;-\;
    X_{v,t}\!\left(\hat{r}_v
      + \sum_{u \in \mathcal{V}} \hat{B}_{vu}\, X_{u,t}\right)
  \right)^{\!2}.
\end{equation}
The reconstructed abundance profile is used for equation \ref{eq:phys-loss} and PyTorch's automatic differentiation engine \cite{pytorch} is used for $\dot{X}_{v,t}$ calculation.
The data loss $\mathcal{L}_{\mathrm{Data}}(\phi)$ combines mean squared reconstruction error on the abundance profiles with a Bray-Curtis dissimilarity (BCD) penalty,
\begin{equation}\label{eq:data-loss}
  \mathcal{L}_{\mathrm{Data}}(\phi)
  \;=\;
  \frac{1}{nT}
  \sum_{v \in \mathcal{V}}\sum_{t=1}^{T}
  \bigl(\hat{X}_{v,t} - X_{v,t}\bigr)^2
  \;+\;
  \frac{\delta}{T}
  \sum_{t=1}^{T}
  \mathrm{BCD}\!\left(\hat{X}_{\cdot,t},\, X_{\cdot,t}\right),
\end{equation}
where $\hat{X}_{v,t}$ is the reconstructed abundance of taxon $v$ at
time $t$, $\delta \geq 0$ weights the BCD term, and
\begin{equation}\label{eq:bcd-def}
  \mathrm{BCD}\!\left(\hat{X}_{\cdot,t},\, X_{\cdot,t}\right)
  \;=\;
  \frac{
    \displaystyle\sum_{v \in \mathcal{V}}
      \bigl|\hat{X}_{v,t} - X_{v,t}\bigr|
  }{
    \displaystyle\sum_{v \in \mathcal{V}}
      \bigl(\hat{X}_{v,t} + X_{v,t}\bigr)
  }.
\end{equation}
Together, equations~\eqref{eq:phys-loss}--\eqref{eq:bcd-def} give concrete form to $\mathcal{L}_{\mathrm{Physics}}$ and $\mathcal{L}_{\mathrm{Data}}$ in equation~\eqref{eq:ki-apinn}. The estimated gLV parameters $\hat{\bm{\theta}} = (\hat{r},\, \hat{B})$ are the output of $\bm{\theta}_\psi(z)$, where $\psi$ collects all trainable weights of $\mathrm{A}^2\mathrm{G}^2$ and
$z = H^{(L)}$ is the knowledge-inclusive representation produced by the encoder-GNN pipeline.


\subsection{KI-APINN for other domains}

\begin{figure}[h]
\centering
\includegraphics[width=0.85\textwidth]{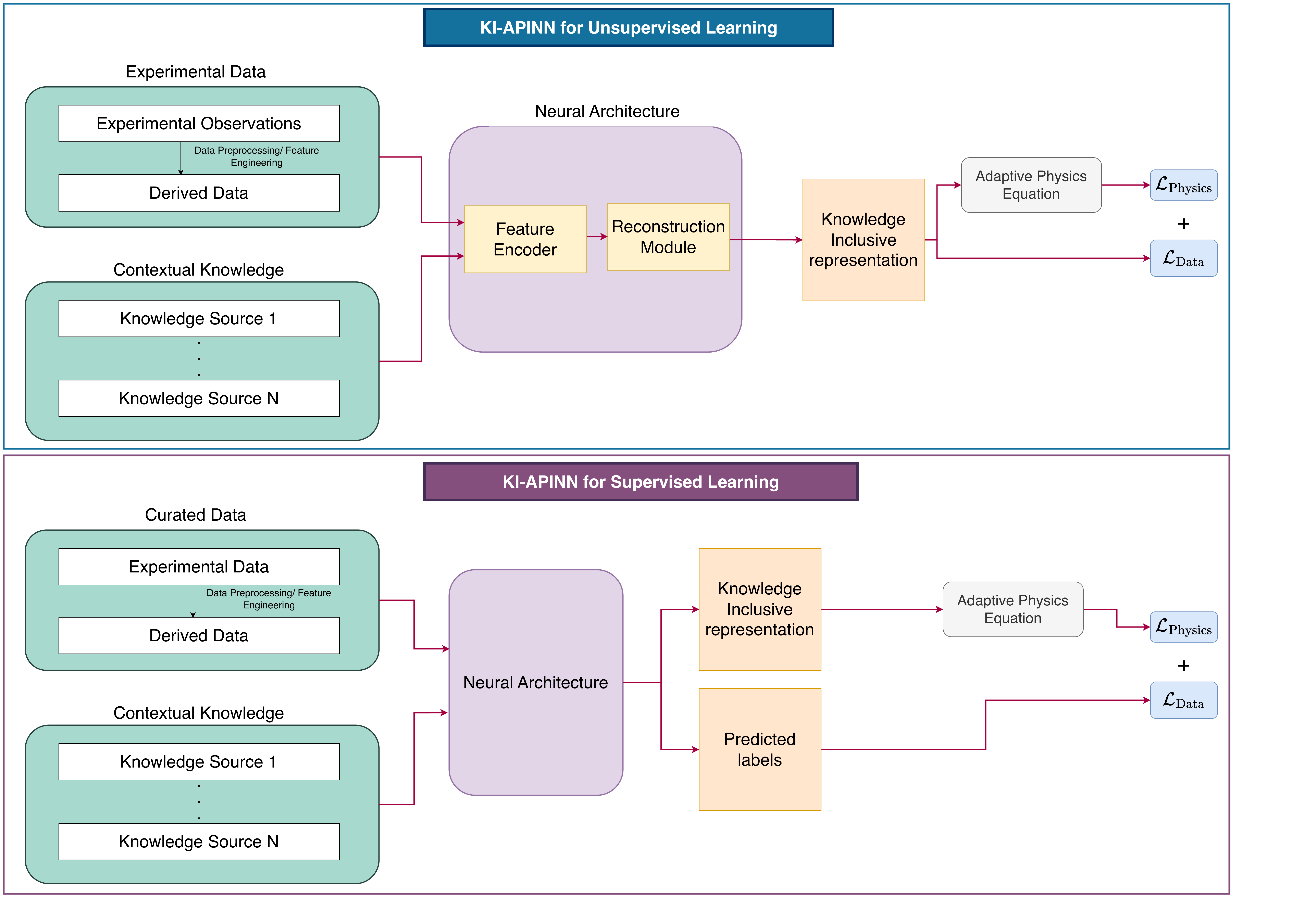}
\caption{The KI-APINN architecture supports both supervised and unsupervised learning settings. In both cases, the parameters of the adaptive physics equation are learned jointly from experimental data and contextual knowledge sources. For unsupervised settings, the underlying neural architecture can include a feature encoder and a reconstruction module. reconstruction data loss can guide the parameter learning process. For supervised settings, a task-specific data loss can be computed from available labels to replace the reconstruction loss.}
\label{fig:ki_apinn_versions}
\end{figure}

The model-agnostic nature of KI-APINN enables application to both supervised and unsupervised settings, as illustrated in Fig. \ref{fig:ki_apinn_versions}. Both settings utilise the adaptive physics equation for the physics loss; the key difference lies in the availability of labels and the corresponding data loss formulation. In unsupervised settings, the neural architecture comprises a feature encoding module and a reconstruction module that produces knowledge-informed representations, with reconstruction error serving as the data loss. In supervised settings, task-specific labels replace reconstruction, providing a supervised data loss.

\subsection{Pipelines and evaluation}

\subsubsection{Mono-community and multi-community pipelines}

\begin{figure}[h]
\centering
\includegraphics[width=1.0\textwidth]{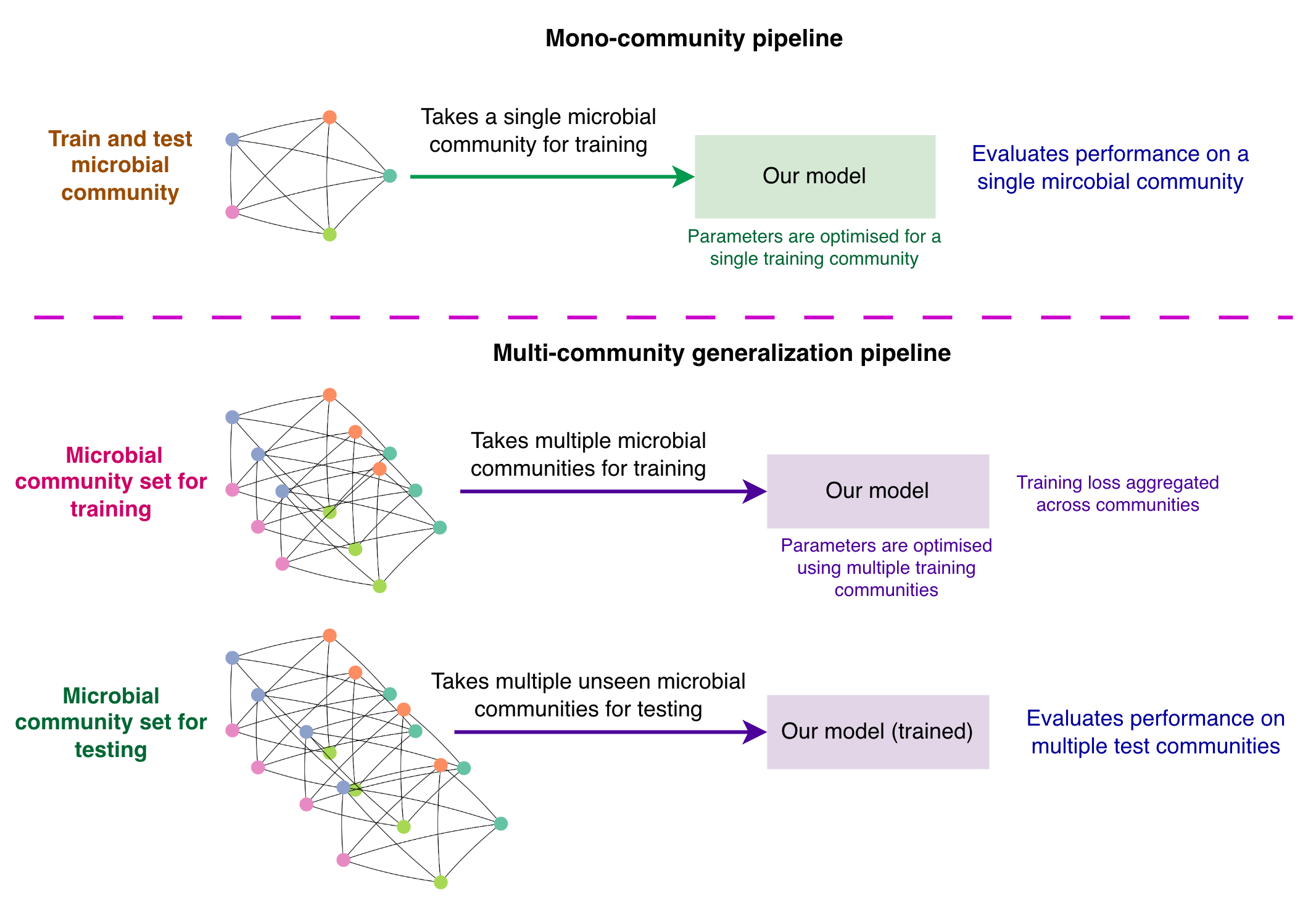}
\caption{Comparison between mono-community and multi-community generalisation pipelines for training and testing. In the mono-community pipeline (top), the model is trained on a single microbial community, with its parameters optimised specifically for that community. The trained model is then evaluated (tested) on the same community, providing insight into its performance within a single-community context. In contrast, the multi-community generalisation pipeline (bottom) trains the model on multiple communities simultaneously, aggregating the training loss across all communities to optimise the model parameters. The trained model is then tested on multiple previously unseen test microbial communities. This comparison highlights the difference between fitting a model to a single community versus learning shared dynamics that can generalise across diverse microbial ecosystems.} 
\label{fig:two_pipelines}
\end{figure}

Two versions of the KI-APINN for microbial interaction modelling have been considered in our experiments: (a) mono-community pipeline and (b) multi-community generalisation pipeline (Fig. \ref{fig:two_pipelines}).

\subsubsection{Evaluation metrics}

To evaluate our pipelines, we use both the BCD-based accuracy and the $\text{R}^2$ score. For the mono-community pipeline, the BCD-based accuracy and the $\text{R}^2$ score are calculated as shown in Eq. \ref{eq:bcd_acc} and \ref{eq:r2_single}.

\begin{align}
\text{BCD Accuracy} &= 
1 - \frac{\sum_{k=0}^{T} BCD(\mathbf{x}_{(t_k)}, \mathbf{x}^*_{(t_k)})}{T}
\label{eq:bcd_acc}
\end{align}

\begin{align}
R^2 = 1 - 
\frac{\sum_{i=1}^{N} \left( x_i - \hat{x}_i \right)^2}
     {\sum_{i=1}^{N} \left( x_i - \bar{x} \right)^2},
\label{eq:r2_single}
\end{align}

In the $\text{R}^2$ calculation, $x_i$ and $\hat{x}_i$ represent the true and reconstructed abundance values, respectively, after flattening the abundance matrices across all microbes and time points. Here, $\bar{x}$ denotes the mean of the true abundance values. An $\text{R}^2$ value of $1$ indicates perfect reconstruction, while $R^2 \leq 0$ implies that the model performs worse than random.

For the multi-community generalisation pipeline, the BCD accuracy is calculated by using Eq. \ref{eq:bcd_acc} for each community and averaging over all the communities. For this pipeline, we have considered three ways of calculating the $\text{R}^2$ score: $\text{R}^2$*: global $\text{R}^2$, $\text{R}^2$†: community-centered $\text{R}^2$, $\text{R}^2$‡: Pearson $\text{R}^2$. $\text{R}^2$* is calculated by averaging the values obtained by using Eq. \ref{eq:r2_single} on each community. For $\text{R}^2$†, the data for each community is centered around their respective community means and used to calculate Eq. \ref{eq:r2_single} on each community and later averaged \cite{ruaud2024modelling}. 


\subsubsection{Threshold selection for performance evaluation}

For a given top $k$ abundant taxa, the cumulative abundance threshold and relative abundance threshold were calculated for the ease of explaining model performance.\\

\noindent\textbf{\textit{Cumulative abundance coverage threshold calculation}}: \\
All taxa were first ranked by their mean abundance across time points. For each sample, the sum of abundances for these top $k$ taxa was computed and divided by the total abundance in that sample, yielding a per-sample coverage fraction. The cumulative abundance coverage threshold was defined as the mean of these per-sample coverage fractions across all samples. This metric quantifies the proportion of total microbial abundance captured by the $k$ most prevalent taxa.\\

\noindent\textbf{\textit{Relative abundance threshold calculation}}: \\
All taxa were first ranked by their mean abundance across time points. For each sample, abundances were then normalised to relative abundances (proportions summing to 1), and the minimum relative abundance among these $k$ taxa was recorded. This process was repeated for all samples, and the minimum relative abundance threshold was defined as the mean of these per-sample minimum values. This metric represents the abundance floor for inclusion in the top $k$ taxa, averaged across all samples.\\

\subsection{Analysis and validation}

Once the microbial community is modeled, the network can be easily obtained to observe different parameters: the pairwise microbial interaction strengths, growth rates for each microbe, and attention scores given to each input source considered for the model. The final microbial network can be easily visualised to observe these interactions in a user-friendly manner. By observing these parameters, we can interpret the roles of individual microbes, quantify the balance between cooperative and competitive interactions, evaluate how these relationships influence community structure and stability, and assess the reliability of each input source in driving model predictions. 

In our modeled network, the top microbial interactions can be validated using MetagenomicKG \cite{metagenomickg}. For each modeled community, the top $k$ interactions can be obtained based on their absolute interaction strengths to get the most influential sub-community. Microbes present in this sub-community are then queried in MetagenomicKG to identify corresponding evidence, specifically known microbial interactions documented within the knowledge graph. Currently, we are considering the interaction type `disease' to obtain microbes that are linked with each other through different diseases.

\backmatter

\section*{Data availability}
All datasets used in this study are publicly available.
Adult gut microbiome datasets (M3 and F4) are publicly available at MG-RAST:4457768.3-4459735.3.
Ginger rhizosphere data can be found at NCBI under BioProject accession number: PRJNA826673.
Neonatal gut and respiratory data can be found in \url{https://github.com/amcdavid/CoordinatedMicrobiome.git}.
\textit{In vitro} gut microbial communities and the code to produce simulated gut microbial communities are available in \url{https://gitlab.eecs.umich.edu/mayank.baranwal/Microbiome.git}.
MetagenomicKG was obtained from \url{https://github.com/KoslickiLab/MetagenomicKG}.
PubMed literature was accessed via PubMed. 
No restrictions apply to data availability.

\section*{Author contributions}
Conceptualisation: S.H, R.R, R.V, S.S, S.L.T; 
Model development and evaluation analyses: R.R, R.V, S.S, A.H, S.H; 
Method development: R.R, R.V, A.H, S.H; 
Implementation: R.R, A.H; 
Metagenomics expertise: S.L.T; 
Paper writing: R.R, A.H, R.V, S.H

\bibliography{sn-bibliography}

\section*{Competing interests}
The authors declare no competing interests.

\section*{Acknowledgements}
We thank all the individuals for participating in this research. This research was supported by the Australian Research Council Discovery Grant (DP210101135). R.R was supported by the Melbourne Research Scholarship.

\noindent

\bigskip

\end{document}